\pdfoutput=1
\PassOptionsToPackage{unicode}{hyperref}
\PassOptionsToPackage{hyphens}{url}
\PassOptionsToPackage{dvipsnames,svgnames,x11names}{xcolor}
\documentclass[
  english,
  12pt,
  letterpaper,
  12pt]{article}
\usepackage{xcolor}
\usepackage[margin=1in]{geometry}
\usepackage{amsmath,amssymb}
\usepackage{iftex}
\ifPDFTeX
  \usepackage[T1]{fontenc}
  \usepackage[utf8]{inputenc}
  \usepackage{textcomp} % provide euro and other symbols
\else % if luatex or xetex
  \usepackage{unicode-math} % this also loads fontspec
  \defaultfontfeatures{Scale=MatchLowercase}
  \defaultfontfeatures[\rmfamily]{Ligatures=TeX,Scale=1}
\fi
\usepackage{lmodern}
\ifPDFTeX
  \usepackage{newtxtext} % Times-like text font for pdfLaTeX (arXiv)
\else
\fi
\IfFileExists{upquote.sty}{\usepackage{upquote}}{}
\IfFileExists{microtype.sty}{% use microtype if available
  \usepackage[]{microtype}
  \UseMicrotypeSet[protrusion]{basicmath} % disable protrusion for tt fonts
}{}
\makeatletter
\ifx\paragraph\undefined\else
  \let\oldparagraph\paragraph
  \renewcommand{\paragraph}{
    \@ifstar
      \xxxParagraphStar
      \xxxParagraphNoStar
  }
  \newcommand{\xxxParagraphStar}[1]{\oldparagraph*{#1}\mbox{}}
  \newcommand{\xxxParagraphNoStar}[1]{\oldparagraph{#1}\mbox{}}
\fi
\ifx\subparagraph\undefined\else
  \let\oldsubparagraph\subparagraph
  \renewcommand{\subparagraph}{
    \@ifstar
      \xxxSubParagraphStar
      \xxxSubParagraphNoStar
  }
  \newcommand{\xxxSubParagraphStar}[1]{\oldsubparagraph*{#1}\mbox{}}
  \newcommand{\xxxSubParagraphNoStar}[1]{\oldsubparagraph{#1}\mbox{}}
\fi
\makeatother

\usepackage{longtable,booktabs,array}
\usepackage{calc} % for calculating minipage widths
\usepackage{etoolbox}
\makeatletter
\patchcmd\longtable{\par}{\if@noskipsec\mbox{}\fi\par}{}{}
\makeatother
\IfFileExists{footnotehyper.sty}{\usepackage{footnotehyper}}{\usepackage{footnote}}
\makesavenoteenv{longtable}
\usepackage{graphicx}
\makeatletter
\newsavebox\pandoc@box
\newcommand*\pandocbounded[1]{% scales image to fit in text height/width
  \sbox\pandoc@box{#1}%
  \Gscale@div\@tempa{\textheight}{\dimexpr\ht\pandoc@box+\dp\pandoc@box\relax}%
  \Gscale@div\@tempb{\linewidth}{\wd\pandoc@box}%
  \ifdim\@tempb\p@<\@tempa\p@\let\@tempa\@tempb\fi% select the smaller of both
  \ifdim\@tempa\p@<\p@\scalebox{\@tempa}{\usebox\pandoc@box}%
  \else\usebox{\pandoc@box}%
  \fi%
}
\def\fps@figure{htbp}
\makeatother

\NewDocumentCommand\citeproctext{}{}
\NewDocumentCommand\citeproc{mm}{%
  \begingroup\def\citeproctext{#2}\cite{#1}\endgroup}
\makeatletter
 \let\@cite@ofmt\@firstofone
 \def\@biblabel#1{}
 \def\@cite#1#2{{#1\if@tempswa , #2\fi}}
\makeatother
\newlength{\cslhangindent}
\newlength{\csllabelwidth}
\newenvironment{CSLReferences}[2] % #1 hanging-indent, #2 entry-spacing
 {\begin{list}{}{%
  \setlength{\itemindent}{0pt}
  \setlength{\leftmargin}{0pt}
  \setlength{\parsep}{0pt}
  \ifodd #1
   \setlength{\leftmargin}{\cslhangindent}
   \setlength{\itemindent}{-1\cslhangindent}
  \fi
  \setlength{\itemsep}{#2\baselineskip}}}
 {\end{list}}
\usepackage{calc}

\ifLuaTeX
\usepackage[bidi=basic,shorthands=off]{babel}
\else
\usepackage[bidi=default,shorthands=off]{babel}
\fi
\ifPDFTeX
\else
\babelfont{rm}[]{Times New Roman}
\fi
\ifLuaTeX
  \usepackage{selnolig} % disable illegal ligatures
\fi

\definecolor{apalink}{HTML}{0074D9}
\definecolor{apatext}{HTML}{000000}

\usepackage{setspace}
\usepackage{fancyhdr}
\usepackage{titlesec}
\usepackage{ragged2e}
\usepackage{etoolbox}
\usepackage{orcidlink}

\usepackage{float}

\RaggedRight

\newcommand{\apashorttitle}{}
\newcommand{\setapashorttitle}[1]{\renewcommand{\apashorttitle}{\MakeUppercase{#1}}}

\newcommand{\apaheadsize}{\fontsize{8}{10}\selectfont}

\makeatletter
\newcommand{\apadotfill}{%
  \leavevmode\cleaders\hb@xt@ .3em{\hss.\hss}\hfill\kern\z@}
\makeatother

\fancypagestyle{apafirstpage}{%
  \fancyhf{}%
  }

\fancypagestyle{apajoufirstpage}{%
  \fancyhf{}%
  \fancyfoot[C]{{\apaheadsize\thepage}}}

\newlength{\apajounarrowwidth}
\newsavebox{\apajouimpactbox}

\usepackage{stfloats}
\usepackage{environ}
\newlength{\apajounoteindent}
\newsavebox{\apajounotebox}

\newcommand{\apajounotetext}{%
  \fontsize{9}{11}\selectfont
  \setlength{\JustifyingParindent}{\apajounoteindent}%
  \justifying
  \setlength{\parindent}{\apajounoteindent}}

\newcommand{\apajounoterule}{%
  \noindent\rule{\linewidth}{0.5pt}%
  \par\vspace{\dimexpr0.4em+3pt\relax}}

\NewEnviron{apajounote}{%
  \sbox{\apajounotebox}{%
    \begin{minipage}{\columnwidth}%
      \apajounotetext\BODY
    \end{minipage}}%
  \ifdim\dimexpr\ht\apajounotebox+\dp\apajounotebox\relax>\dimexpr\textheight/3\relax
    \begin{figure*}[b]%
      \apajounoterule
      \apajounotetext\BODY
    \end{figure*}%
  \else
    \begin{figure}[b]%
      \apajounoterule
      \usebox{\apajounotebox}%
    \end{figure}%
  \fi}

\makeatletter
\renewcommand*\l@figure{\@dottedtocline{1}{0em}{0em}}
\let \l@table \l@figure
\makeatother

\makeatletter
\renewcommand*\l@section{\@dottedtocline{1}{0em}{0em}}
\renewcommand*\l@subsection{\@dottedtocline{2}{1.5em}{0em}}
\renewcommand*\l@subsubsection{\@dottedtocline{3}{3em}{0em}}
\makeatother

\newlength{\apamastheadrulewidth}
\newlength{\apalogoheight}
\newsavebox{\apamastheadbox}

\newlength{\apajouparindent}
\makeatletter
\def\@BLTrule{\@@BLTrule()}
\makeatother

\makeatletter
\newtoks\apa@lt@foot
\long\def\apa@lt@firsthead#1\endhead{\apa@lt@head}
\def\apa@lt@head{\apa@lt@grab}
\long\def\apa@lt@grab#1\endlastfoot{\noalign{\global\apa@lt@foot{#1}}}
\newcommand{\apalongtableastabular}{%
  \renewenvironment{longtable}[2][]{%
    \global\apa@lt@foot{}%
    \let\endfirsthead\apa@lt@firsthead
    \let\endhead\apa@lt@head
    \let\endfoot\relax
    \let\endlastfoot\relax
    \begin{tabular}{##2}%
  }{%
    \the\apa@lt@foot
    \end{tabular}%
  }}
\makeatother

\titleformat{\section}[block]
  {\normalfont\bfseries\centering}{}{0pt}{}
\titlespacing{\section}{0pt}{\baselineskip}{0pt}

\titleformat{\subsection}[block]
  {\normalfont\bfseries\raggedright}{}{0pt}{}
\titlespacing{\subsection}{0pt}{\baselineskip}{0pt}

\titleformat{\subsubsection}[block]
  {\normalfont\bfseries\itshape\raggedright}{}{0pt}{}
\titlespacing{\subsubsection}{0pt}{\baselineskip}{0pt}

\titleformat{\paragraph}[runin]
  {\normalfont\bfseries}{}{0pt}{}
\titlespacing{\paragraph}{\parindent}{\baselineskip}{0.5em}

\titleformat{\subparagraph}[runin]
  {\normalfont\bfseries\itshape}{}{0pt}{}
\titlespacing{\subparagraph}{\parindent}{\baselineskip}{0.5em}

\newcommand{\apatitle}[1]{%
  {\centering\bfseries #1\par}}

\newenvironment{apaauthor}
  {\par\centering}
  {\par}

\newcommand{\apatitlepageheading}[1]{%
  {\centering\bfseries #1\par}}

\newenvironment{apanoindent}
  {\par\setlength{\parindent}{0pt}}
  {\par}

\newcommand{\apafloattitle}[1]{%
  {\par\noindent\bfseries #1\par}}
\newcommand{\apafloatcaption}[1]{%
  {\par\noindent\itshape #1\par}}

\newcommand{\apafigureskip}{\vspace{1.2em}}
\newenvironment{apafloatnote}
  {\par\setlength{\parindent}{0pt}\noindent}
  {\par}

\newlength{\apatablenotegap}
\newcommand{\apatablenoteskip}{\par\vspace{\apatablenotegap}}

\makeatletter
\@ifundefined{cslhangindent}{}{\setlength{\cslhangindent}{0.5in}}
\makeatother

\newlength{\apathesisentrysep}
\newcommand{\apathesissingle}{%
  \setlength{\apathesisentrysep}{\baselineskip}%
  \setstretch{1}\selectfont
  \addtolength{\apathesisentrysep}{-\baselineskip}}

\makeatletter
\newcommand{\apathesisreferences}{%
  \@ifundefined{CSLReferences}{}{%
    \apptocmd{\CSLReferences}%
      {\apathesissingle
       \setlength{\itemsep}{\apathesisentrysep}%
       \advance\@outerparskip\apathesisentrysep}{}{}}}
\makeatother

\makeatletter
\newcommand{\apathesisnotes}{%
  \renewcommand\@makefntext[1]{%
    \setlength{\parindent}{0.5in}%
    \indent\@makefnmark\,##1}%
  {\apathesissingle\footnotesize
   \global\footnotesep=\dimexpr 2\baselineskip - \dp\strutbox\relax}}
\makeatother

\makeatletter
\newcommand{\apajouhangindent}{%
  \@ifundefined{cslhangindent}{}{\setlength{\cslhangindent}{\apajouparindent}}}
\makeatother
\usepackage{booktabs}
\usepackage{array}
\usepackage{float}
\usepackage{bm}
\usepackage{amsmath}
\usepackage{placeins}
\usepackage{tcolorbox}
\tcbuselibrary{listings,skins,breakable}
\usepackage[skip=10pt]{caption}
\newtcblisting{cotexample}{listing only, breakable, enhanced jigsaw, colback=gray!6, colframe=gray!35, boxrule=0.4pt, arc=1pt, left=5pt, right=5pt, top=3pt, bottom=3pt, boxsep=1pt, before skip=0.75\baselineskip, after skip=0.75\baselineskip, listing options={basicstyle=\ttfamily\normalsize, breaklines=true, columns=fullflexible, aboveskip=0pt, belowskip=0pt, lineskip=-1pt}}
\makeatletter
\@ifpackageloaded{float}{}{\usepackage{float}}
\floatstyle{plain}
\@ifundefined{c@chapter}{\newfloat{illustration}{h}{loill}}{\newfloat{illustration}{h}{loill}[chapter]}
\floatname{illustration}{Illustration}

\makeatother
\makeatletter
\@ifpackageloaded{caption}{}{\usepackage{caption}}
\AtBeginDocument{%
\ifdefined\contentsname
  \renewcommand*\contentsname{Table of contents}
\else
  \newcommand\contentsname{Table of contents}
\fi
\ifdefined\listfigurename
  \renewcommand*\listfigurename{List of Figures}
\else
  \newcommand\listfigurename{List of Figures}
\fi
\ifdefined\listtablename
  \renewcommand*\listtablename{List of Tables}
\else
  \newcommand\listtablename{List of Tables}
\fi
\ifdefined\figurename
  \renewcommand*\figurename{Figure}
\else
  \newcommand\figurename{Figure}
\fi
\ifdefined\tablename
  \renewcommand*\tablename{Table}
\else
  \newcommand\tablename{Table}
\fi
}
\@ifpackageloaded{float}{}{\usepackage{float}}
\floatstyle{ruled}
\@ifundefined{c@chapter}{\newfloat{codelisting}{h}{lop}}{\newfloat{codelisting}{h}{lop}[chapter]}
\floatname{codelisting}{Listing}

\makeatother
\makeatletter
\@ifpackageloaded{caption}{}{\usepackage{caption}}
\@ifpackageloaded{subcaption}{}{\usepackage{subcaption}}
\makeatother

\setapashorttitle{LLM COT FOR Q-MATRIX}
\usepackage{bookmark}
\IfFileExists{xurl.sty}{\usepackage{xurl}}{} % add URL line breaks if available
\hypersetup{
  pdflang={en},
  colorlinks=true,
  linkcolor={apalink},
  filecolor={apalink},
  citecolor={apalink},
  urlcolor={apalink},
  pdfcreator={LaTeX via pandoc}}

\author{Jihong Zhang \and Zijing Jin \and Mingfeng Xue}
\date{}
\begin{document}

\hfill\break

\hfill\break

\vspace*{3\baselineskip}

\apatitle{

Reasoning-Trace Analysis Using Large Language Models for Interpretable
Q-Matrix Construction

}

\begin{apaauthor}

\hfill\break

Jihong Zhang\textsuperscript{1}, Zijing Jin\textsuperscript{2}, and Mingfeng Xue\textsuperscript{3}

\textsuperscript{1}Department of Counseling, Leadership, and Research
Methods, University of Arkansas

\textsuperscript{2}School of Computing and Information, University of
Pittsburgh

\textsuperscript{3}Department of Information, Library, and Research
Sciences, University of North Carolina at Greensboro

\end{apaauthor}

\hfill\break

\hfill\break

\apatitlepageheading{

Author Note

}

Jihong Zhang \orcidlink{0000-0003-2820-3734}
\url{https://orcid.org/0000-0003-2820-3734}

Zijing Jin \orcidlink{0009-0003-4036-2559}
\url{https://orcid.org/0009-0003-4036-2559}

Correspondence concerning this article should be addressed to Jihong
Zhang, Department of Counseling, Leadership, and Research
Methods, University of Arkansas, 109 Graduate Education
Building, Fayetteville, AR 72703, Email:
\href{mailto:jzhang@uark.edu}{\nolinkurl{jzhang@uark.edu}}

\clearpage

\apatitlepageheading{

Abstract

}

\begin{apanoindent}

Large language models (LLMs) can generate Q-matrices for diagnostic
classification models that approximate expert specifications, but
evaluations have focused on the final matrices, leaving the basis of
individual item--attribute assignments unexamined. This opacity limits
the responsible use of AI in measurement. We propose Reasoning-Trace
Analysis (RTA), a framework in which an LLM generates attribute-labeled
reasoning steps through chain-of-thought (CoT) prompting and these steps
are analyzed to reconstruct, compare, and inspect each assignment. RTA
is intended to supplement expert review and empirical validation. In a
simulation study with perturbed fraction subtraction items, four LLMs
recovered the required attributes and withheld attributes the items did
not require. In an empirical study with 11 fraction subtraction items,
responses from 536 examinees, four LLMs, up to three temperatures, and
100 replications per condition, CoT-derived Q-matrices overlapped
substantially with the reference Q-matrix (mean overlap .809) but less
than directly generated Q-matrices (.877), and they varied more across
replications. Inspection of the traces linked disagreements to item-type
coding, labeling rules, and how operations were expressed as steps.
Q-matrices with similar overlap differed in fit to examinee responses.
RTA provides inspectable evidence that matrix-level agreement and fit do
not.

\end{apanoindent}


\clearpage

\apatitle{

Reasoning-Trace Analysis Using Large Language Models for Interpretable
Q-Matrix Construction

}

Diagnostic classification models (DCMs), also called cognitive
diagnostic models (CDMs), are confirmatory latent class models that
classify examinees into discrete mastery profiles over predefined
attributes. The validity and reliability of their classifications depend
on a well-specified Q-matrix that links each item to the attributes it
is intended to measure. Q-matrices are typically specified through
expert judgment, which is effortful and largely subjective, and then
verified and revised with examinee responses and other evidence. Large
language models (LLMs) offer a potential way to construct Q-matrices
more efficiently, and recent studies suggest that LLMs acting as experts
can generate Q-matrices that approximate human expert references and,
under some conditions, achieve competitive model fit
(\citeproc{ref-asiretEvaluatingChatGPT4osPerformance2025}{Aşiret \& Ömür
Sünbül, 2025};
\citeproc{ref-xueEvaluatingGeneralPurposeMultimodal2026}{Xue \&
Appleton, 2026}; \citeproc{ref-zhangCDMQMatrixDiscovery2026}{Zhang \&
Rao, 2026}). Whether a Q-matrix is specified by experts or generated by
an LLM, evaluating its validity requires access to the reasoning behind
individual item--attribute assignments. Yet expert reasoning is rarely
documented in an inspectable form, and evaluations of LLM-generated
Q-matrices mostly focus on the final matrices and their empirical
performance. Where a rationale is requested, it accompanies the Q-vector
rather than producing it
(\citeproc{ref-zhangCDMQMatrixDiscovery2026}{Zhang \& Rao, 2026}). In
both cases, potential misspecifications remain difficult to detect,
replicate, or contest. For LLM-generated Q-matrices, this opacity poses
a threat to the responsible use of AI in measurement, which calls for
AI-supported decisions to be transparent, explainable, and accountable
(\citeproc{ref-johnsonResponsibleAIMeasurement2025}{Johnson, 2025}).

Chain-of-thought (CoT) prompting can elicit intermediate reasoning steps
before a final answer (\citeproc{ref-weiChainThoughtPrompting2022}{Wei
et al., 2022}). In Q-matrix construction, these traces provide
observable material for examining how a model connects an item's
solution process to proposed cognitive attributes. A related precedent
is the application of Schoenfeld's Episode Theory, originally developed
for human mathematical problem-solving, to the analysis of reasoning
traces from large reasoning models
(\citeproc{ref-liUnderstandingThinkingProcess2025}{M. Li, Zhang, et al.,
2025}). However, stated reasoning may not reflect the factors driving a
model's predictions
(\citeproc{ref-lanhamMeasuringFaithfulnessChain2023}{Lanham et al.,
2023}; \citeproc{ref-turpinLanguageModelsDont2023}{Turpin et al.,
2023}). Thus, whether elicited traces can provide a sound basis for
evaluating the reasoning underlying individual item--attribute
assignments remains an open question.

To address this question, we propose Reasoning-Trace Analysis (RTA), a
framework that treats an elicited trace as a claim to be examined
against external criteria rather than as direct evidence of the model's
reasoning. Here, we apply RTA to Q-matrix construction. In RTA, an LLM
first generates reasoning steps labeled with the attributes they
exercise through CoT prompting, and these labeled steps are then
analyzed to reconstruct and examine each item--attribute assignment.
This extends Q-matrix evaluation beyond the final matrix to the support
for individual item--attribute assignments. RTA is intended to
supplement existing Q-matrix specification by providing an initial
Q-matrix with an inspectable basis for each assignment that remains
subject to expert review and empirical validation
(\citeproc{ref-johnsonResponsibleAIMeasurement2025}{Johnson, 2025}). We
evaluate the performance and usefulness of RTA through a simulation
study and an empirical study. The following sections review the relevant
background, describe RTA, and state the research questions before
reporting the two studies.

\section{Background}\label{background}

\subsection{Q-Matrix Construction by Human Experts and
LLMs}\label{q-matrix-construction-by-human-experts-and-llms}

DCMs support fine-grained skill profiling in domains ranging from
mathematics (\citeproc{ref-chenRegularizedLatentClass2017c}{Chen et al.,
2017}) and reading
(\citeproc{ref-liConstructingValidatingQMatrix2013}{H. Li \& Suen,
2013}) to clinical and affective constructs
(\citeproc{ref-izaLatentStructureSocial2014}{Iza et al., 2014};
\citeproc{ref-ruppDiagnosticMeasurementTheory2010}{Rupp et al., 2010}).
Q-matrix construction in DCMs requires analysts to define the relevant
attributes, justify item--attribute links, and evaluate the proposed
matrix against substantive and empirical evidence
(\citeproc{ref-buck1998}{Buck et al., 1998};
\citeproc{ref-tatsuokaRuleSpaceApproach1983b}{Tatsuoka, 1983}). Existing
procedures often rely on think-aloud protocols, expert ratings,
committee deliberation, and response-data validation
(\citeproc{ref-liConstructingValidatingQMatrix2013}{H. Li \& Suen,
2013}). These procedures remain labor-intensive, time-consuming, and
costly, and misspecified entries can distort item parameter estimates
and examinee classification
(\citeproc{ref-ruppEffectsQMatrixMisspecification2008}{Rupp \& Templin,
2008}).

When response data are available, empirical validation methods can
identify item--attribute assignments whose modification may improve fit.
Two such methods are the proportion of variance accounted for (PVAF)
method (\citeproc{ref-delaTorreChiuGeneralMethod2016}{de la Torre \&
Chiu, 2016}) and the Hull method
(\citeproc{ref-najeraBalancingFitParsimony2021}{Nájera et al., 2021}).
Both de la Torre and Chiu
(\citeproc{ref-delaTorreChiuGeneralMethod2016}{2016}) and Culpepper
(\citeproc{ref-culpepperEstimatingCognitiveDiagnosis2019}{2019})
illustrated their methods with the fraction subtraction data from which
the present items are drawn. These methods require response data, and
their suggestions still call for expert review before adoption
(\citeproc{ref-najeraBalancingFitParsimony2021}{Nájera et al., 2021}).

LLM applications in assessment include scoring open-ended responses
(\citeproc{ref-leeApplyingLargeLanguage2024}{Lee et al., 2024};
\citeproc{ref-mendoncaEvaluatingLLMsAutomated2025}{Mendonça et al.,
2025}), generating items
(\citeproc{ref-chanAutomaticItemGeneration2025}{Chan et al., 2025}),
modeling item difficulty (\citeproc{ref-liItemDifficultyModeling2025}{M.
Li, Jiao, et al., 2025}), and producing synthetic respondents for item
calibration (\citeproc{ref-liuLeveragingLLMRespondents2025b}{Liu et al.,
2025}). Recent LLM-based Q-matrix studies suggest that LLMs can
approximate matrices built by human experts at the overall matrix level.
Aşiret and Ömür Sünbül
(\citeproc{ref-asiretEvaluatingChatGPT4osPerformance2025}{2025}) showed
that ChatGPT-4o can draft Q-matrices directly that match human expert
references on most entries. They evaluated the generated Q-matrices
using attribute-level and attribute-vector overlap rates, empirical
Q-matrix validation, Q-matrix recovery rates, and CDM model-fit indices.
The study by Xue and Appleton
(\citeproc{ref-xueEvaluatingGeneralPurposeMultimodal2026}{2026}) showed
that AI-generated Q-matrices can approximate human expert baselines with
competitive model fit. Their evaluation used chatbot--human matching
accuracy, Q-matrix density, classification discrepancies, and model-fit
statistics across model and prompting conditions. Fan et al.
(\citeproc{ref-fanUseAIToolsQMatrix2026}{2026}) compared Q-matrices
generated by several general-purpose LLMs for a reading comprehension
test with the validated Q-matrix of H. Li and Suen
(\citeproc{ref-liConstructingValidatingQMatrix2013}{2013}) and found
substantial variation in agreement across models, with lower agreement
for newer model versions. Zhang and Rao
(\citeproc{ref-zhangCDMQMatrixDiscovery2026}{2026}) combined an
LLM-proposed Q-matrix with response-based refinement under the
deterministic inputs, noisy ``and'' gate (DINA) model and asked the LLM
for a brief rationale for each Q-vector, which experts then rated.

The metrics used in these studies show whether the final generated
Q-matrix resembles a human expert matrix or yields acceptable model-data
fit. However, they do not show how each item--attribute assignment is
inferred or whether a disagreement reflects model error, prompt
sensitivity, or an alternative problem-solving path. From a measurement
perspective, validity evidence based on the response processes behind
these matrices is therefore missing, which limits the responsible use of
AI in measurement. Moreover, the expert reference is itself a
specification rather than a ground truth, so fit to examinee responses
must be examined alongside agreement with it. In addition, repeated
sampling of the same prompt yields different traces and sometimes
different answers, so aggregation across generations has become a common
way to summarize what a model produces
(\citeproc{ref-wangSelfConsistencyImprovesChain2022}{Wang et al.,
2023}). For Q-matrix construction, Kim
(\citeproc{ref-kimHowReproducibleAre2026}{2026}) found that regenerating
a Q-matrix with the same model at the same temperature changed the
mastery profiles assigned to many examinees, and that majority voting
across runs did not remove this instability.

\subsection{Responsible AI in
Assessment}\label{responsible-ai-in-assessment}

Johnson (\citeproc{ref-johnsonResponsibleAIMeasurement2025}{2025})
described responsible AI for measurement as the use of AI to improve how
human capabilities are measured while protecting individual rights.
Among its principles are transparency, explainability, and
accountability, together with the requirement that inferences made by AI
be valid for their intended use. The same report places final authority
over AI-supported judgments with human educators, an approach described
as human-in-the-loop. Bulut et al.
(\citeproc{ref-bulutRiseArtificialIntelligence2024}{2024}) similarly
identified validity, reliability, transparency, and fairness among the
central ethical concerns raised by AI in educational measurement. In
Q-matrix construction, an LLM acting as an expert makes decisions that
determine what a diagnostic assessment measures. If the basis of these
decisions cannot be inspected, the validity of the resulting Q-matrix
cannot be fully examined, and human reviewers have little on which to
base their judgment.

\subsection{Reasoning Traces From
LLMs}\label{reasoning-traces-from-llms}

One directly observable form of evidence for LLM reasoning is CoT
output. CoT produces observable intermediate reasoning steps, either by
supplying worked examples
(\citeproc{ref-weiChainThoughtPrompting2022}{Wei et al., 2022}) or by
instructing the model to reason step by step without examples
(\citeproc{ref-kojimaLargeLanguageModels2022}{Kojima et al., 2022}).
Compared with direct prompting, CoT often provides modest average gains
in accuracy and can produce less stable answers in some settings
(\citeproc{ref-meinckeDecreasingValueChain2025}{Meincke et al., 2025}).

LLM-generated reasoning can be conceptualized as a sequence of
problem-solving episodes rather than as a single final output. Drawing
on Schoenfeld's Episode Theory, which was originally developed to
characterize human mathematical problem solving, M. Li, Zhang, et al.
(\citeproc{ref-liUnderstandingThinkingProcess2025}{2025}) analyzed the
output of reasoning models using six episode types: Read, Analyze, Plan,
Implement, Explore, and Verify. Their analysis indicated that elicited
reasoning traces can exhibit episode-level structures that resemble
categories used to describe human problem solving. This correspondence
does not establish that LLMs reason like humans, but it provides a
conceptual basis for examining how generated steps align with
human-defined cognitive skills.

Beyond classifying the forms of reasoning expressed in model output,
prior work has examined the mechanisms by which transformer models solve
mathematical problems. Hu et al.
(\citeproc{ref-huCaseBasedRuleBased2024}{2024}) distinguished case-based
reasoning from rule-based reasoning. In their framework, case-based
reasoning solves a new problem by retrieving or matching similar cases
encountered during training, whereas rule-based reasoning applies an
abstract rule to new problems. Across intervention experiments on five
mathematical tasks, the authors found that the tested transformer models
relied primarily on case-based reasoning, even when scratchpads were
used. This finding indicates that a correct answer may be reached
without applying the intended rule.

Consequently, an elicited reasoning trace is not, by itself, evidence
that the stated reasoning is correct, faithful, or aligned with the
intended cognitive attributes. The extent to which a final answer
depends on the stated steps varies across tasks and model sizes
(\citeproc{ref-lanhamMeasuringFaithfulnessChain2023}{Lanham et al.,
2023}), and some apparent reasoning is decorative rather than
load-bearing (\citeproc{ref-zhaoCanAhaMoments2025}{Zhao et al., 2025}).
It therefore remains unclear whether elicited traces recover the
attributes an item requires, even when those attributes are known in
advance.

\subsection{Verbal Reports and Response-Process
Evidence}\label{verbal-reports-and-response-process-evidence}

RTA draws its logic from the analysis of human verbal reports. Protocol
analysis treats concurrent verbal reports as data on the cognitive
processes used in a task and specifies conditions under which such
reports are valid and complete
(\citeproc{ref-ericssonProtocolAnalysisVerbal1993}{Ericsson \& Simon,
1993}). In achievement testing, what an examinee says while solving an
item is coded against external criteria rather than accepted as a
transparent record of cognition
(\citeproc{ref-leightonAvoidingMisconceptionMisuse2004}{Leighton,
2004}), and Leighton
(\citeproc{ref-leightonUsingThinkAloudInterviews2017}{2017})
distinguished think-aloud interviews for problem solving from cognitive
laboratory interviews for comprehension. The \emph{Standards for
Educational and Psychological Testing}
(\citeproc{ref-aeraStandardsEducationalPsychological2014}{American
Educational Research Association et al., 2014}) list response processes
as a source of validity evidence, and Padilla and Benítez
(\citeproc{ref-padillaValidityEvidenceBased2014}{2014}) reviewed the
methods available for obtaining it. These works concern human examinees.
RTA applies the same logic to the reasoning steps of an LLM.

\section{Overview of Reasoning-Trace
Analysis}\label{overview-of-reasoning-trace-analysis}

This section describes the proposed RTA framework, which will be
evaluated in the simulation and empirical studies. RTA consists of two
components: trace generation (Step 0), in which an LLM produces
attribute-labeled reasoning traces through CoT prompting, and a
four-step analysis of these traces against external criteria (see
\hyperref[fig-rta-framework]{Figure~\ref*{fig-rta-framework}}). The
external criteria are a set of expert-defined attributes and a reference
Q-matrix that specifies the attributes each item is expected to require.

\begin{figure}[H]

\setcounter{figure}{0}

\refstepcounter{figure}%
\makeatletter\def\@currentlabel{1}\makeatother%
\label{fig-rta-framework}

\apafloattitle{Figure 1}

\apafloatcaption{Reasoning-Trace Analysis Framework}

\apafigureskip

\addcontentsline{lof}{figure}{Figure 1. Reasoning-Trace Analysis
Framework}

\includegraphics[width=1\linewidth,height=\textheight,keepaspectratio]{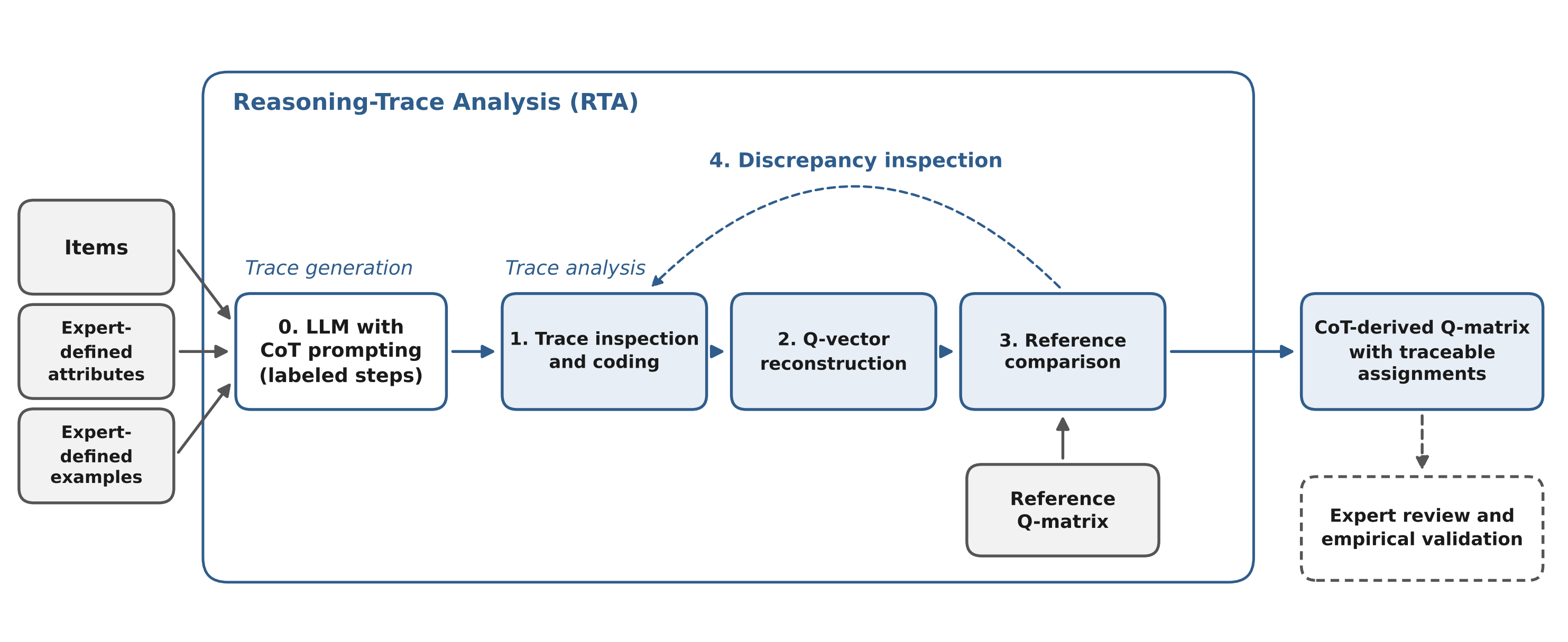}

\begin{apafloatnote}

\emph{Note}. Expert-defined~examples are optional. Expert review and
empirical validation (dashed box) lie outside RTA.

\end{apafloatnote}

\end{figure}

\subsection{Trace Generation}\label{trace-generation}

Trace generation is inspired by the rule-following principle of
Rule-Following Fine-Tuning (RFFT), proposed by Hu et al.
(\citeproc{ref-huCaseBasedRuleBased2024}{2024}). RFFT trains transformer
models to recite and apply explicit rules step by step and improved
length generalization in their addition experiments. RTA adapts this
principle at inference time without fine-tuning the model. Given an
item, the expert-defined attributes, and optionally expert-defined
examples, the LLM is instructed to solve the item step by step and label
each step with the attribute it exercises, using the format
\texttt{Step\ N\ {[}SKILL{]}:\ description}. Procedural steps are
labeled \texttt{{[}Other{]}}, and the response ends with
\texttt{Answer:\ \textless{}value\textgreater{}}. The expert-defined
examples can be worked examples of the labeled format
(\citeproc{ref-weiChainThoughtPrompting2022}{Wei et al., 2022}), or they
can be omitted and the model instructed to reason step by step
(\citeproc{ref-kojimaLargeLanguageModels2022}{Kojima et al., 2022}).

\subsection{Trace Analysis}\label{trace-analysis}

The labeled reasoning steps are then analyzed in four steps. First, in
\emph{trace inspection and coding}, the stepwise outputs are inspected
against the expert-defined attributes to identify attributes that are
used, omitted, merely mentioned, or introduced despite not being
required by the item's solution. Second, in \emph{Q-vector
reconstruction}, the Q-vector of each item is reconstructed from the
model-generated skill labels. An attribute is marked present if at least
one reasoning step is labeled with it, and the resulting matrix is
referred to as the \emph{CoT-derived Q-matrix}. Third, in
\emph{reference comparison}, the CoT-derived Q-matrix is compared with
the reference Q-matrix to identify item--attribute disagreements.
Finally, in \emph{discrepancy inspection}, these disagreements are
traced to the corresponding reasoning steps to examine whether they
involve omitted or misclassified attributes or potentially legitimate
alternative solution paths. Consistent with the supplementary role of
RTA described in the Introduction, the resulting CoT-derived Q-matrix
serves as an initial specification for expert review and empirical
validation.

\section{Current Study}\label{current-study}

The current study evaluates RTA for Q-matrix construction using fraction
subtraction as an example, a domain with an established expert reference
Q-matrix used in cognitive diagnosis research
(\citeproc{ref-tatsuoka1990}{Tatsuoka, 1990}). The evaluation is guided
by three open issues that remain to be addressed, which are stated below
as research questions. Both studies used four expert-defined attributes.
A1 (basic fraction subtraction) involves subtracting numerators after
equalizing denominators. A2 (simplifying/reducing) involves reducing
fractions to lowest terms or eliminating zero components. A3 (separating
whole number from fraction) involves decomposing mixed numbers or
converting improper fractions. A4 (borrowing from whole number) involves
converting one whole unit into a fraction when the minuend fraction is
smaller than the subtrahend fraction.

\subsection{Workflows}\label{workflows}

Two workflows were compared (see
\hyperref[fig-workflows]{Figure~\ref*{fig-workflows}}). The CoT tracing
workflow (hereafter, the CoT workflow) implements RTA as described in
the Overview. The empirical study used CoT prompting without worked
examples, whereas the simulation study used one worked example. The
prompt below shows the version used in the empirical study. The full
wording of the simulation study's prompt is provided in the
Supplementary Materials (see Section S1.1).

\begin{cotexample}
Skills:
- A1: basic fraction subtraction
- A2: simplifying/reducing
- A3: separating whole number
- A4: borrowing from whole number

Format each step as: Step N [SKILL]: description
End with Answer: <value>.

Rules:
- Label a skill only if the step actually performs that operation.
- If a step only checks, states, or mentions an operation without performing it, label it [Other].
- If a result is already in simplest form, label that step [Other] rather than [A2].
- Use multiple skill labels separated by & when a single step performs more than one operation.
- Use [Other] for procedural steps such as writing the final answer.

Solve step by step.

Problem: 4 4/12 - 2 7/12
\end{cotexample}

The direct Q-matrix workflow (hereafter, the direct workflow)
corresponds to the direct generation examined in prior studies and
served as a baseline for matrix-level comparison in the empirical study.
The LLM was given the item content and short labels for the same four
attributes and asked to generate the full Q-matrix in a single response.
No stepwise reasoning or skill labels were requested, so this workflow
did not provide traces for step-level analysis.

\begin{figure}[H]

\setcounter{figure}{1}

\refstepcounter{figure}%
\makeatletter\def\@currentlabel{2}\makeatother%
\label{fig-workflows}

\apafloattitle{Figure 2}

\apafloatcaption{CoT and Direct Q-Matrix Workflows}

\apafigureskip

\addcontentsline{lof}{figure}{Figure 2. CoT and Direct Q-Matrix
Workflows}

\includegraphics[width=1\linewidth,height=\textheight,keepaspectratio]{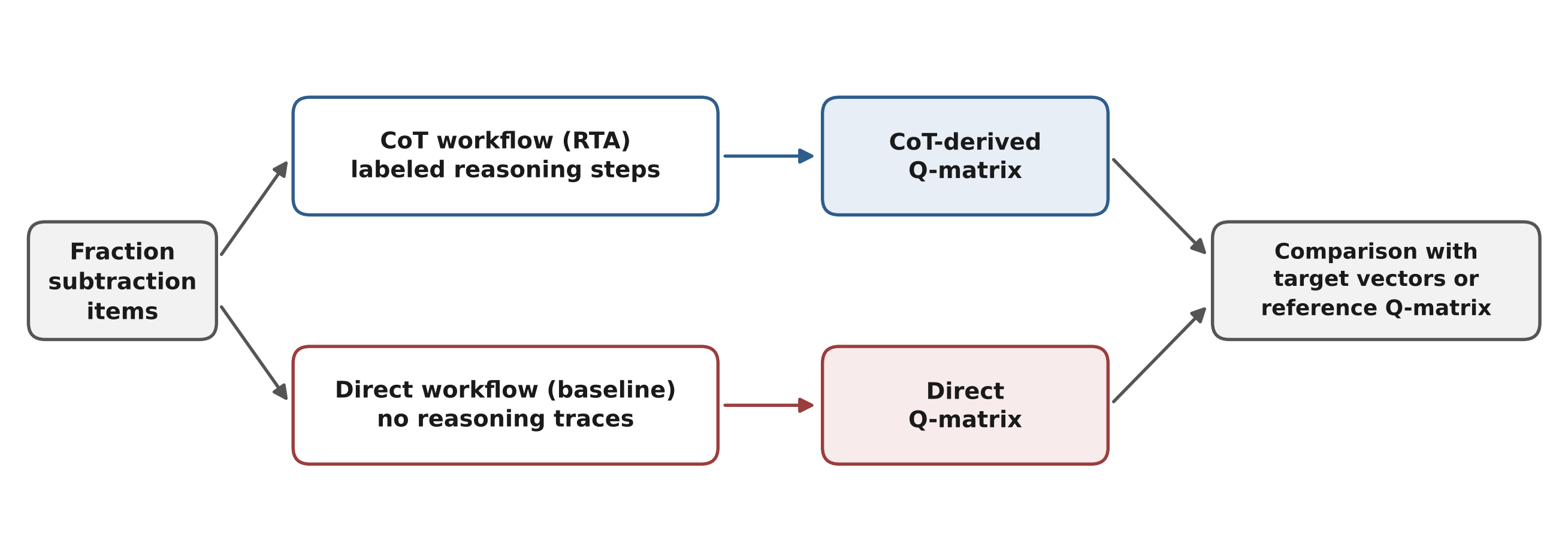}

\begin{apafloatnote}

\emph{Note}. The~direct workflow was used only in the empirical study.

\end{apafloatnote}

\end{figure}

\subsection{Research Questions and
Hypotheses}\label{research-questions-and-hypotheses}

The study addressed three research questions. \emph{RQ1
(Reasoning-Process Recovery):} To what extent do reasoning steps
generated through the CoT workflow recover the expected item--attribute
reasoning process? \emph{RQ2 (LLM--Human Alignment):} How closely do
generated Q-matrices align with the reference Q-matrix, where and in
which direction do they diverge, how do they compare in fit to examinee
responses, and do these patterns differ between the two workflows?
\emph{RQ3 (Q-Matrix Stability and Variation):} How stable are generated
Q-matrices across repeated runs, how much do they vary across models and
temperature settings, and do these patterns differ between the two
workflows? RQ1 is a precondition for the other two, because a
CoT-derived Q-matrix is only as informative as the traces from which it
is reconstructed. The simulation study addresses RQ1 by testing trace
recovery under controlled conditions and provides a benchmark for
interpreting the empirical study, which addresses RQ2 and RQ3.

Three working hypotheses guided the study. \emph{H1 (Reasoning-Process
Recovery):} The CoT workflow would recover the intended item--attribute
reasoning process when item design specified the required attributes in
advance. Although elicited traces can resemble categories of human
problem solving (\citeproc{ref-liUnderstandingThinkingProcess2025}{M.
Li, Zhang, et al., 2025}), stated reasoning does not necessarily reflect
the factors driving model responses
(\citeproc{ref-lanhamMeasuringFaithfulnessChain2023}{Lanham et al.,
2023}; \citeproc{ref-turpinLanguageModelsDont2023}{Turpin et al.,
2023}). Support for H1 would indicate that the traces contain the
information needed to construct a CoT-derived Q-matrix. \emph{H2
(LLM--Human Alignment):} CoT-derived Q-matrices would meaningfully
overlap with the reference Q-matrix, with an overlap magnitude
comparable to that of the direct workflow. This expectation draws on
findings that LLM-generated Q-matrices approximate expert references
(\citeproc{ref-asiretEvaluatingChatGPT4osPerformance2025}{Aşiret \& Ömür
Sünbül, 2025};
\citeproc{ref-xueEvaluatingGeneralPurposeMultimodal2026}{Xue \&
Appleton, 2026}) and that CoT offers modest average gains over direct
prompting (\citeproc{ref-meinckeDecreasingValueChain2025}{Meincke et
al., 2025}). Disagreement with the reference indicates where expert
review is needed. \emph{H3 (Q-Matrix Stability and Variation):}
Generated Q-matrices would vary across runs, models, and temperature
settings, reflecting sensitivity to generation conditions and
potentially different solution paths. Repeated generation with the same
model can alter examinees' assigned mastery profiles
(\citeproc{ref-kimHowReproducibleAre2026}{Kim, 2026}), and we expected
greater run-to-run variability under the CoT workflow than under the
direct workflow, because CoT can produce less stable answers than direct
prompting (\citeproc{ref-meinckeDecreasingValueChain2025}{Meincke et
al., 2025}). Stability determines whether generated Q-matrices support
reproducible classification.

\subsection{Models and Consensus
Aggregation}\label{models-and-consensus-aggregation}

Both studies used four LLMs selected after preliminary analysis and
API-feasibility pilot testing: GPT-4o mini, DeepSeek v4 Flash, Claude
Haiku 4.5, and Kimi K2.6. Temperature controlled the sampling randomness
of the LLMs, with lower values producing more deterministic outputs and
higher values allowing more varied outputs. Its permitted range differs
by provider, from 0 to 1 or from 0 to 2 for the models used here. The
model and API configurations, together with the settings for all
conditions, are provided in the Supplementary Materials (see Section
S1.2).

Following the aggregation approach noted in the Background
(\citeproc{ref-wangSelfConsistencyImprovesChain2022}{Wang et al.,
2023}), replicated outputs were aggregated to summarize item-level
attribute assignments. Where a design included replications, an
attribute was included in the consensus Q-matrix if at least half of the
replications assigned that attribute to the item.

\section{Simulation Study}\label{simulation-study}

To address RQ1, we examined whether the reasoning traces generated in
RTA recover the required attributes, using items whose required
attributes were known in advance. Because each item's target vector, the
Q-vector implied by its design, was known, the Q-vector reconstructed
from each trace could be checked directly against it.

\subsection{Data}\label{data}

The simulation study used 10 perturbed versions of the item
\(4\frac{4}{12} - 2\frac{7}{12}\) from Tatsuoka's 20-item,
eight-attribute fraction subtraction item set
(\citeproc{ref-tatsuoka1990}{Tatsuoka, 1990}). We selected this item
because its intended solution requires all four attributes (A1 to A4)
and yields a longer stepwise solution with more opportunities for skill
labeling. One standard solution process for this item is as follows:

\[
4\frac{4}{12} - 2\frac{7}{12}
\xrightarrow{(\text{A4})} \boxed{3\frac{16}{12} - 2\frac{7}{12}}
\xrightarrow{(\text{A3},\text{A1})} \boxed{1\frac{9}{12}}
\xrightarrow{(\text{A2})} \boxed{1\frac{3}{4}}.
\]

From this base item, we created 10 perturbed items. Each perturbed item
changed exactly one numerical element while preserving the requirement
for all four attributes. For the two denominator perturbations, the
shared denominator was changed in both fractions.
\hyperref[tbl-perturbation]{Table~\ref*{tbl-perturbation}} presents the
10 perturbed items, the numerical element changed, the expected
solution, and the expected answer.

\par\addvspace{\baselineskip}

\setcounter{table}{0}

\refstepcounter{table}%
\makeatletter\def\@currentlabel{1}\makeatother%
\label{tbl-perturbation}

\apafloattitle{Table 1}

\apafloatcaption{Perturbed Items}

\addcontentsline{lot}{table}{Table 1. Perturbed Items}

{\def\LTcaptype{none} % do not increment counter
\begin{longtable}[]{@{}
  >{\centering\arraybackslash}p{(\linewidth - 8\tabcolsep) * \real{0.0600}}
  >{\raggedright\arraybackslash}p{(\linewidth - 8\tabcolsep) * \real{0.2000}}
  >{\raggedright\arraybackslash}p{(\linewidth - 8\tabcolsep) * \real{0.1800}}
  >{\raggedright\arraybackslash}p{(\linewidth - 8\tabcolsep) * \real{0.4000}}
  >{\centering\arraybackslash}p{(\linewidth - 8\tabcolsep) * \real{0.1600}}@{}}
\toprule\noalign{}
\begin{minipage}[b]{\linewidth}\centering
No.
\end{minipage} & \begin{minipage}[b]{\linewidth}\raggedright
Element
\end{minipage} & \begin{minipage}[b]{\linewidth}\raggedright
Item
\end{minipage} & \begin{minipage}[b]{\linewidth}\raggedright
Solution
\end{minipage} & \begin{minipage}[b]{\linewidth}\centering
Answer
\end{minipage} \\
\midrule\noalign{}
\endhead
\bottomrule\noalign{}
\endlastfoot
1 & Whole1 +1 & \(5\frac{4}{12} - 2\frac{7}{12}\) &
\(4\frac{16}{12} - 2\frac{7}{12} = 2\frac{9}{12} = 2\frac{3}{4}\) &
\(2\frac{3}{4}\) \\
2 & Whole1 -1 & \(3\frac{4}{12} - 2\frac{7}{12}\) &
\(2\frac{16}{12} - 2\frac{7}{12} = \frac{9}{12} = \frac{3}{4}\) &
\(\frac{3}{4}\) \\
3 & Whole2 +1 & \(4\frac{4}{12} - 3\frac{7}{12}\) &
\(3\frac{16}{12} - 3\frac{7}{12} = \frac{9}{12} = \frac{3}{4}\) &
\(\frac{3}{4}\) \\
4 & Whole2 -1 & \(4\frac{4}{12} - 1\frac{7}{12}\) &
\(3\frac{16}{12} - 1\frac{7}{12} = 2\frac{9}{12} = 2\frac{3}{4}\) &
\(2\frac{3}{4}\) \\
5 & Num1 -1 & \(4\frac{3}{12} - 2\frac{7}{12}\) &
\(3\frac{15}{12} - 2\frac{7}{12} = 1\frac{8}{12} = 1\frac{2}{3}\) &
\(1\frac{2}{3}\) \\
6 & Num1 +1 & \(4\frac{5}{12} - 2\frac{7}{12}\) &
\(3\frac{17}{12} - 2\frac{7}{12} = 1\frac{10}{12} = 1\frac{5}{6}\) &
\(1\frac{5}{6}\) \\
7 & Num2 -1 & \(4\frac{4}{12} - 2\frac{6}{12}\) &
\(3\frac{16}{12} - 2\frac{6}{12} = 1\frac{10}{12} = 1\frac{5}{6}\) &
\(1\frac{5}{6}\) \\
8 & Num2 +1 & \(4\frac{4}{12} - 2\frac{8}{12}\) &
\(3\frac{16}{12} - 2\frac{8}{12} = 1\frac{8}{12} = 1\frac{2}{3}\) &
\(1\frac{2}{3}\) \\
9 & Denom 12 to 15 & \(4\frac{4}{15} - 2\frac{7}{15}\) &
\(3\frac{19}{15} - 2\frac{7}{15} = 1\frac{12}{15} = 1\frac{4}{5}\) &
\(1\frac{4}{5}\) \\
10 & Denom 12 to 18 & \(4\frac{4}{18} - 2\frac{7}{18}\) &
\(3\frac{22}{18} - 2\frac{7}{18} = 1\frac{15}{18} = 1\frac{5}{6}\) &
\(1\frac{5}{6}\) \\
\end{longtable}
}

\apatablenoteskip

\begin{apafloatnote}

\emph{Note}. Each~perturbed item changes one numerical element of the
base item 4 4/12 - 2 7/12 and requires all four attributes. A Denom
change applies to both fractions.

\end{apafloatnote}

\par\addvspace{\baselineskip}

We also constructed six variants of the base item in which at least one
attribute was absent from the target vector. These variants modified the
required operations by eliminating borrowing, making fraction reduction
unnecessary, removing the whole-number parts, or using an integer
subtrahend. For example, \(4\frac{9}{12} - 2\frac{3}{12}\) requires no
borrowing, so its target vector is \([1, 1, 1, 0]\). Target vectors
followed the item-classification convention of the reference Q-matrix.
For \(4\frac{4}{12} - 2\), this convention codes A1 as present, as the
reference does for the structurally identical item \(3\frac{7}{8} - 2\),
whereas an operation-level coding would code it as absent. The six
variants and their target vectors are listed in the Supplementary
Materials (see Table S1).

\subsection{Design}\label{design}

The perturbed items tested whether the CoT workflow recovered the
required attributes. They varied surface numerical features while
requiring all four attributes, so that any alternative problem-solving
path a model took could be controlled. The simulation prompt defined
each attribute with examples of when it should and should not be used.
Each of the four LLMs solved the perturbed items individually under the
CoT workflow, with one generation per item. GPT-4o mini, DeepSeek v4
Flash, and Claude Haiku 4.5 were run at temperature 0, whereas Kimi K2.6
was run at temperature 0.6, the only value its endpoint accepted when
thinking was disabled. Each generated Q-vector was compared with the
target vector \([1, 1, 1, 1]\).

Because every perturbed item required all four attributes, a model that
always returned \([1, 1, 1, 1]\) would match every target regardless of
item content. The variants therefore tested whether the CoT workflow
withheld attributes an item did not require. Each of the four models
generated 10 replications per variant at the temperature used for the
perturbed items. For each variant, the consensus vector---the
corresponding row of the consensus Q-matrix---was compared with its
target vector. The variants were run later through a different API
route, described in the Supplementary Materials (see Section S1.2).

\subsection{Evaluation Criteria}\label{evaluation-criteria}

We evaluated the recovery of the Q-vectors reconstructed from the
generated traces, together with the traces themselves, using three
criteria at different levels: \emph{overlap} of item--attribute
assignments, \emph{accuracy} of final answers, and \emph{variation} in
reasoning steps. Overlap measures agreement between generated and target
item--attribute assignments at the entry, attribute, and item levels.
Each disagreement corresponds to an entry that would be misspecified if
the reconstructed Q-vector were adopted, with the consequences of
misspecification described in the Background. Accuracy measures the
proportion of final answers that match the expected answers (see
\hyperref[tbl-perturbation]{Table~\ref*{tbl-perturbation}} and Table S1)
and serves as an outcome-level check separate from attribute agreement.
Because a trace that reaches an incorrect answer may omit or
misrepresent required operations, accuracy shows whether attribute
recovery depends on solving the item correctly. Variation refers to
differences in the labeled reasoning sequences that different models
produced for the same item, including step ordering and the omission of
expected attributes, and is referred to as \emph{step-level variation}.
It shows whether models that recover the same attributes do so through
the same solution path.

For the perturbed items, an attribute counted as recovered if at least
one reasoning step was labeled with it. For the variants, recovery was
assessed using the consensus vector. Step order was examined
descriptively as step-level variation, not as a recovery criterion,
because different sequences can express the same set of required
attributes. When a reconstructed Q-vector differed from its target, we
applied discrepancy inspection to the corresponding trace.

\subsubsection{Overlap}\label{overlap}

We report three overlap indices: the overall overlap rate, the
attribute-level overlap rate, and the exact item match rate. Let
\(q_{ik}\) denote the generated assignment for attribute \(k\) in item
\(i\), coded as 1 if the attribute was assigned according to the rule
above and 0 otherwise. Let \(t_{ik}\) denote the corresponding target
assignment, coded as 1 if the item requires attribute \(k\) and 0
otherwise. Let \(\mathbf{q}_i\) and \(\mathbf{t}_i\) denote the
generated and target Q-vectors of item \(i\). Let \(J\) be the number of
items evaluated and \(K=4\) the number of attributes. The indicator
function \(I(\cdot)\) equals 1 when its condition is true and 0
otherwise.

The \emph{overall overlap rate}, equivalent to the Q-matrix recovery
rate of Nájera et al.
(\citeproc{ref-najeraBalancingFitParsimony2021}{2021}), is the
proportion of all item--attribute assignments that match the target:

\[
O=\frac{1}{JK}\sum_{i=1}^{J}\sum_{k=1}^{K}I(q_{ik}=t_{ik}).
\]

The \emph{attribute-level overlap rate} for attribute \(k\) is the
proportion of items for which the generated assignment matches the
target for that attribute:

\[
O_k=\frac{1}{J}\sum_{i=1}^{J}I(q_{ik}=t_{ik}).
\]

The \emph{exact item match rate} (EIMR) is the proportion of items for
which the generated Q-vector matches the target vector on all four
attributes:

\[
\text{EIMR}=\frac{1}{J}\sum_{i=1}^{J}I(\mathbf{q}_i=\mathbf{t}_i).
\]

For the perturbed items, all target entries were 1, so the
attribute-level overlap rate \(O_k\) equals the true positive rate for
attribute \(k\). For the variants, overall overlap was interpreted
relative to an all-ones baseline, representing a model that always
assigned all four attributes regardless of item content.

\subsection{Results}\label{results}

\subsubsection{Perturbed Items}\label{perturbed-items}

DeepSeek v4 Flash, Claude Haiku 4.5, and Kimi K2.6 solved all 10
perturbed items correctly and recovered the full target vector for every
item (\(O = \text{EIMR} = 1.00\)). Accuracy and attribute recovery
showed different patterns for GPT-4o mini. It solved 5 of the 10 items
correctly (accuracy = .50) but recovered the full four-attribute vector
for 8 of the 10 items (\(\text{EIMR} = .80\), \(O = .95\)). Its
attribute-level overlap rates were 1.00 for A1 and A2 and .90 for A3 and
A4. These findings suggest that labeling the required operations and
solving the item rely on different processes for GPT-4o mini.

\par\addvspace{\baselineskip}

\setcounter{table}{1}

\refstepcounter{table}%
\makeatletter\def\@currentlabel{2}\makeatother%
\label{tbl-prelim-example}

\apafloattitle{Table 2}

\apafloatcaption{Example Results From the CoT Workflow for One Perturbed
Item (\(4\frac{5}{12} - 2\frac{7}{12}\))}

\addcontentsline{lot}{table}{Table 2. Example Results From the CoT
Workflow for One Perturbed Item (\(4\frac{5}{12} - 2\frac{7}{12}\))}

{\def\LTcaptype{none} % do not increment counter
\begin{longtable}[]{@{}
  >{\raggedright\arraybackslash}p{(\linewidth - 2\tabcolsep) * \real{0.1900}}
  >{\raggedright\arraybackslash}p{(\linewidth - 2\tabcolsep) * \real{0.8100}}@{}}
\toprule\noalign{}
\begin{minipage}[b]{\linewidth}\raggedright
Model
\end{minipage} & \begin{minipage}[b]{\linewidth}\raggedright
Reasoning-step content
\end{minipage} \\
\midrule\noalign{}
\endhead
\bottomrule\noalign{}
\endlastfoot
GPT-4o mini & Step 1 {[}A1{]}: Subtract the fractions directly since
they have the same denominator (5/12 - 7/12 = -2/12). Step 2
{[}A3\&A4{]}: Since the result of the fraction subtraction is negative,
borrow from the whole number (4 5/12 = 3 17/12). Step 3 {[}A1{]}:
Subtract the whole numbers and fractions (3 - 2 = 1 and 17/12 - 7/12 =
10/12). Step 4 {[}A2{]}: Simplify the fraction (10/12 = 5/6). Step 5
{[}Other{]}: Write the final answer. \\
DeepSeek v4 Flash & Step 1 {[}A4{]}: Borrow one from the whole number (4
5/12 = 3 17/12). Step 2 {[}A3\&A1{]}: Separate the whole numbers and
fractions, then subtract each part (3 - 2 = 1 and 17/12 - 7/12 = 10/12).
Step 3 {[}A2{]}: Simplify the result (1 10/12 -\textgreater{} 1 5/6).
Step 4 {[}Other{]}: Write the final answer. \\
Claude Haiku 4.5 & Step 1 {[}A4{]}: Borrow one from the whole number (4
5/12 = 3 17/12). Step 2 {[}A3\&A1{]}: Separate the whole numbers and
fractions, then subtract each part (3 - 2 = 1 and 17/12 - 7/12 = 10/12).
Step 3 {[}A2{]}: Simplify the fraction (10/12 = 5/6). Step 4
{[}Other{]}: Write the final answer. \\
Kimi K2.6 & Step 1 {[}A4{]}: Borrowing one from the whole number (4 5/12
= 3 17/12). Step 2 {[}A3\&A1{]}: Separate the whole numbers and
fractions, then subtract each part (3 - 2 = 1 and 17/12 - 7/12 = 10/12).
Step 3 {[}A2{]}: Simplify the result (1 10/12 -\textgreater{} 1 5/6).
Step 4 {[}Other{]}: Write the final answer. \\
\end{longtable}
}

\apatablenoteskip

\begin{apafloatnote}

\emph{Note}. Stepwise~text is condensed from the raw output.

\end{apafloatnote}

\par\addvspace{\baselineskip}

\FloatBarrier

\begin{figure}[H]

\setcounter{figure}{2}

\refstepcounter{figure}%
\makeatletter\def\@currentlabel{3}\makeatother%
\label{fig-cot-ribbon}

\apafloattitle{Figure 3}

\apafloatcaption{Tagged CoT Ribbons for Four Perturbed Items}

\apafigureskip

\addcontentsline{lof}{figure}{Figure 3. Tagged CoT Ribbons for Four
Perturbed Items}

\includegraphics[width=1\linewidth,height=\textheight,keepaspectratio]{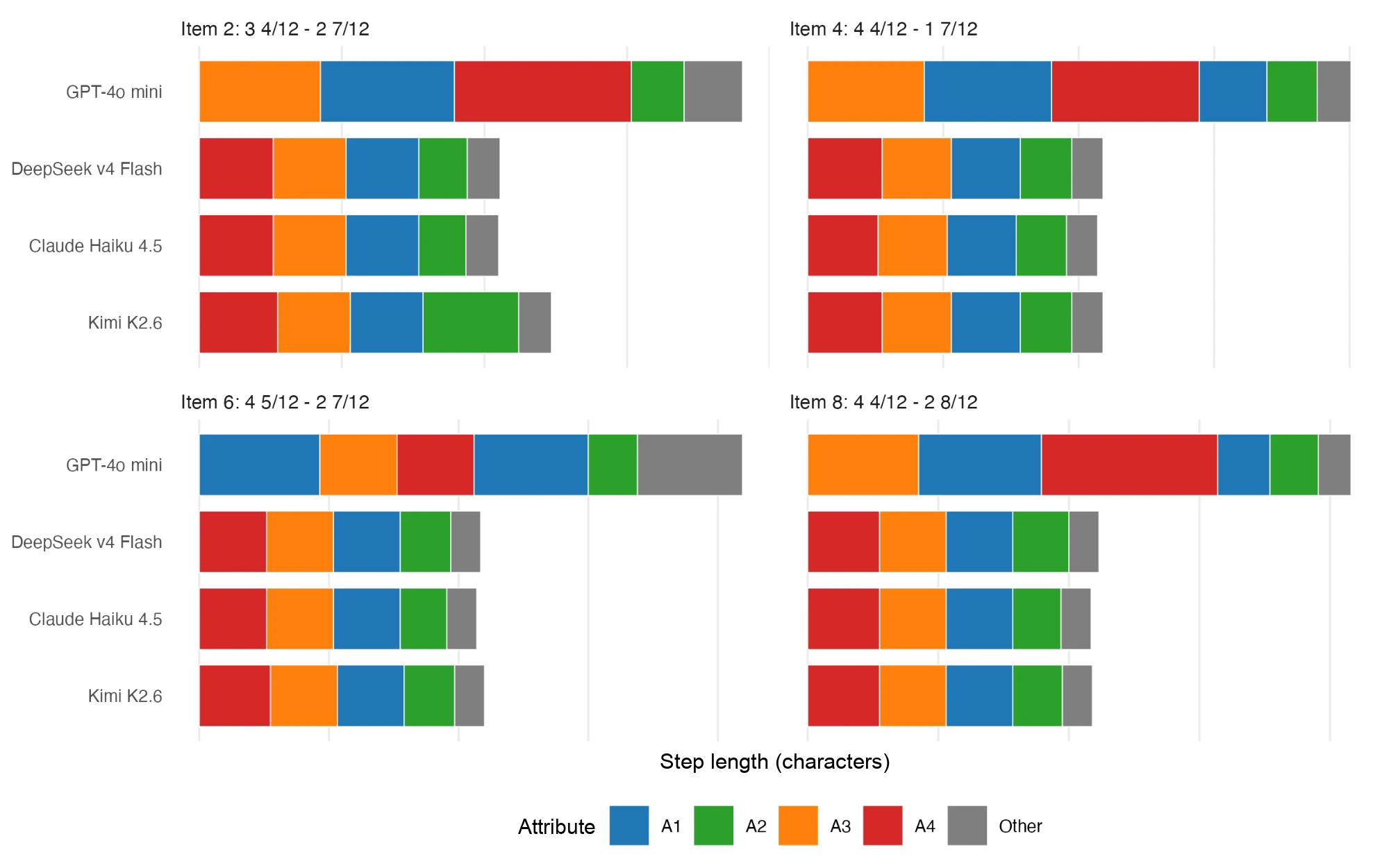}

\begin{apafloatnote}

\emph{Note}. Segment~width is proportional to step length, and color
indicates the labeled attribute. A step with two labels is split into
two equal segments. The remaining six perturbed items are shown in
Figure S1 in the Supplementary Materials.

\end{apafloatnote}

\end{figure}

GPT-4o mini followed a different step order from the other three models
on all 10 perturbed items. It attempted subtraction before borrowing and
introduced borrowing only after an intermediate result became negative.
The other three models consistently borrowed before subtracting and so
avoided a negative intermediate result. Nevertheless, for the example
shown in \hyperref[tbl-prelim-example]{Table~\ref*{tbl-prelim-example}},
all four models represented the same set of attributes. The step-level
variation in \hyperref[fig-cot-ribbon]{Figure~\ref*{fig-cot-ribbon}}
illustrates these differences in step ordering and attribute labeling.
On perturbed Items 5 and 9, GPT-4o mini carried out an expected
operation but folded it into an adjacent step's label (borrowing under
{[}A2{]} on Item 5 and whole-number separation under {[}A1{]} on Item
9), so A4 and A3 were not credited. The other three models produced
closely matching step sequences. Ribbons for four representative items
(perturbed Items 2, 4, 6, and 8) are shown here, and the remaining six
are provided in the Supplementary Materials (see Figure S1).

\subsubsection{Variants With Absent
Attributes}\label{variants-with-absent-attributes}

Across the six variants, the four models' consensus vectors achieved
overall overlap rates ranging from .958 to 1.000, corresponding to 23 or
24 matches among the 24 item--attribute assignments (see Table S1). A
model that always returned \([1, 1, 1, 1]\) would achieve an overlap
rate of .625 (15 of 24 assignments). Across all four models, A4 was
withheld on all five variants that required no borrowing, A3 was
withheld on both variants without whole-number parts, and A2 was
withheld on both variants that required no reduction. Accuracy was .967
for GPT-4o mini and 1.000 for the other three models.

Two discrepancies occurred in the variants. For
\(4\frac{4}{11} - 2\frac{7}{11}\), GPT-4o mini performed whole-number
separation within a step labeled {[}A1{]} in 7 of 10 replications, which
resulted in the omission of A3 from its consensus vector. For
\(4\frac{4}{12} - 2\), DeepSeek v4 Flash omitted A1 in 8 of 10
replications, whereas the consensus vectors of the other three models
assigned A1. Under operation-level coding, A1 was treated as absent for
this item, and overall overlap rates ranged from .917 to 1.000, compared
with .583 for the all-ones baseline. All four models exceeded the
corresponding all-ones baseline under both item-classification and
operation-level coding.

Taken together, the CoT workflow recovered the required attributes under
controlled conditions. Three models recovered every target vector across
the perturbed items, and GPT-4o mini recovered 8 of 10, including items
it answered incorrectly. In the variants, all four models withheld the
attributes that the items did not require and exceeded the all-ones
baseline under both coding conventions. Both types of discrepancy
reflected how operations were labeled. An operation was performed but
labeled with an adjacent attribute, or, for the integer subtrahend, the
label followed operation-level coding. Step order differed across models
without changing the set of attributes recovered.

\section{Empirical Study}\label{empirical-study}

The empirical study addressed RQ2 and RQ3 by applying RTA through the
CoT workflow to fraction subtraction items with an established reference
Q-matrix, using the direct workflow as a baseline.

\subsection{Data}\label{data-1}

The empirical study used 11 items selected from the same fraction
subtraction set (\citeproc{ref-tatsuoka1990}{Tatsuoka, 1990}). The
reference Q-matrix was taken from de la Torre and Chiu
(\citeproc{ref-delaTorreChiuGeneralMethod2016}{2016}) and follows Method
B of Mislevy (\citeproc{ref-mislevyTestTheoryReconceived1996}{1996}),
which separates mixed numbers into whole-number and fractional parts and
borrows from the whole number when necessary. To evaluate model fit, we
used responses from 536 examinees to the same 11 items, distributed with
the GDINA R package (\citeproc{ref-maGDINARPackage2020}{Ma \& de la
Torre, 2020}). The items and reference Q-matrix are provided in the
Supplementary Materials (see Table S2).

\subsection{Design}\label{design-1}

The design crossed model, temperature, and workflow using the same four
LLMs as the simulation study. Kimi K2.6 was evaluated at temperature
0.6, while the other three models were evaluated at temperatures 0, 0.5,
and 1.0. This yielded 20 conditions (see
\hyperref[tbl-empirical-design]{Table~\ref*{tbl-empirical-design}}): 18
from crossing the three models, three temperatures, and two workflows,
plus two Kimi K2.6 conditions, one for each workflow. Each condition
included 100 replications. Within a replication, the CoT workflow solved
the 11 items one at a time, whereas the direct workflow generated the
full \(11 \times 4\) matrix at once. In both workflows, the complete set
of item--attribute assignments from one replication counted as one
generated Q-matrix. The design therefore produced 2,000 generated
Q-matrices (20 conditions × 100 replications).

\par\addvspace{\baselineskip}

\setcounter{table}{2}

\refstepcounter{table}%
\makeatletter\def\@currentlabel{3}\makeatother%
\label{tbl-empirical-design}

\apafloattitle{Table 3}

\apafloatcaption{Empirical Study Experimental Factors}

\addcontentsline{lot}{table}{Table 3. Empirical Study Experimental
Factors}

{\def\LTcaptype{none} % do not increment counter
\begin{longtable}[]{@{}
  >{\raggedright\arraybackslash}p{(\linewidth - 4\tabcolsep) * \real{0.2600}}
  >{\raggedright\arraybackslash}p{(\linewidth - 4\tabcolsep) * \real{0.2600}}
  >{\raggedright\arraybackslash}p{(\linewidth - 4\tabcolsep) * \real{0.4800}}@{}}
\toprule\noalign{}
\begin{minipage}[b]{\linewidth}\raggedright
Factor
\end{minipage} & \begin{minipage}[b]{\linewidth}\raggedright
Level
\end{minipage} & \begin{minipage}[b]{\linewidth}\raggedright
Setting
\end{minipage} \\
\midrule\noalign{}
\endhead
\bottomrule\noalign{}
\endlastfoot
Model & GPT-4o mini, DeepSeek v4 Flash, Claude Haiku 4.5 & Low
reasoning-effort setting where available \\
& Kimi K2.6 & Thinking disabled \\
Temperature & 0, 0.5, 1.0 & Applied to GPT-4o mini, DeepSeek v4 Flash,
and Claude Haiku 4.5 \\
& 0.6 & Applied to Kimi K2.6 \\
Workflow & CoT workflow & Item-level prompt with stepwise skill
labels \\
& Direct workflow & Attribute labels and item content followed by direct
11 by 4 Q-matrix generation \\
\end{longtable}
}

\par\addvspace{\baselineskip}

\subsection{Evaluation Criteria}\label{evaluation-criteria-1}

We evaluated the generated Q-matrices, including the CoT-derived
Q-matrices produced by RTA, using five criteria: \emph{alignment},
\emph{structural identifiability}, \emph{stability}, \emph{accuracy},
and \emph{model fit}. Alignment measures agreement with the reference
Q-matrix. As in the Simulation Study, each disagreement marks a
potentially misspecified entry, and its location shows where a generated
matrix would need expert review. Structural identifiability assesses
whether the matrix supports identifiable parameters of the DINA model. A
Q-matrix that does not meet the identifiability conditions described
below leaves the model parameters not strictly identifiable, however
closely it agrees with the reference. Stability measures consistency
across repeated generations. When a generated Q-matrix changes from one
generation to the next, the resulting classifications depend on which
generation is used. Accuracy measures the correctness of the final
answers and shows whether the traces behind a CoT-derived Q-matrix come
from correct solutions. Model fit evaluates how well the generated
matrices account for examinee responses and provides evidence
independent of the reference Q-matrix, which is itself an expert
specification. Alignment and model fit addressed RQ2, while stability
addressed RQ3. Structural identifiability and accuracy provided
additional evidence for interpreting the results. We examined variation
across models, temperatures, and workflows.

\subsubsection{Alignment}\label{alignment}

For each model--temperature--workflow condition, we estimated the
probability of assigning attribute \(k\) to item \(i\) across the
\(R=100\) replications:

\[
\hat{P}_{ik}=\frac{1}{R}\sum_{r=1}^{R}q_{ikr},
\] where \(q_{ikr}=1\) if attribute \(k\) was assigned to item \(i\) in
replication \(r\), and 0 otherwise. Under the CoT workflow, assignments
were reconstructed from the generated skill labels, whereas under the
direct workflow they were read from the generated matrix.

A consensus Q-matrix was constructed, as described in Models and
Consensus Aggregation, by assigning 1 when \(\hat{P}_{ik}\geq.50\) and 0
otherwise. Each consensus matrix was compared with the reference
Q-matrix using the overall overlap rate, attribute-level overlap rate,
and exact item match rate, as defined in the Simulation Study's
Evaluation Criteria section. With 11 items and four attributes, the
overall overlap rate covers 44 item--attribute assignments. Item-level
overlap was also reported to identify individual items that were
difficult to recover. Its average across items equals the overall
overlap rate.

To characterize disagreement patterns, we counted \emph{omissions}
(attributes present in the reference but absent from the generated
matrix) and \emph{over-labelings} (attributes absent from the reference
but present in the generated matrix), which correspond to the under- and
over-specifications of Nájera et al.
(\citeproc{ref-najeraBalancingFitParsimony2021}{2021}). These counts
were computed for the CoT workflow at temperature 0.5 (0.6 for Kimi
K2.6). We also calculated matrix density, the proportion of the 44 cells
assigned a value of 1. Because thresholding can conceal uncertainty in
assignments with probabilities near .50, we report cell-level assignment
probabilities alongside the aggregate overlap indices where relevant.
For the CoT-derived Q-matrices, we then applied discrepancy inspection
to each disagreement.

\subsubsection{Structural
Identifiability}\label{structural-identifiability}

We assessed whether each of the 2,000 generated Q-matrices and 20
consensus Q-matrices met the necessary and sufficient conditions for
strict identifiability of DINA model parameters
(\citeproc{ref-guSufficientNecessaryCondition2019}{Gu \& Xu, 2019}). The
three conditions were: (1) \emph{completeness}, meaning that the
Q-matrix contains an identity submatrix after row permutation, so each
attribute has at least one single-attribute item; (2) \emph{repeated
measurement}, requiring each attribute to be assessed by at least three
items; and (3) \emph{distinctness}, requiring the columns of the
submatrix remaining after removal of the identity rows to be pairwise
distinct. We applied the same criteria to the reference Q-matrix. A
matrix was classified as structurally identifiable only if all three
conditions were met. This assessment concerned DINA identifiability, not
agreement with the reference Q-matrix.

\subsubsection{Stability}\label{stability}

We assessed Q-matrix stability across the 100 replications within each
condition using two indices. The \emph{dominant Q-matrix proportion},
\(D\), was the proportion of replications yielding the most frequently
generated full Q-matrix:

\[
D=\frac{\max_m n_m}{R},
\] where \(n_m\) is the number of replications that produced matrix
configuration \(m\), and \(R=100\). Higher values indicate greater
concentration on a single matrix.

The \emph{unstable item--attribute assignment rate}, \(U\), was the
proportion of item--attribute assignments that varied across
replications:

\[
U=\frac{1}{JK}\sum_{i=1}^{J}\sum_{k=1}^{K}
I(0<\hat{P}_{ik}<1),
\] where \(J=11\), \(K=4\), \(I(\cdot)\) is the indicator function
defined in the Simulation Study's Evaluation Criteria, and
\(\hat{P}_{ik}\) is the condition-specific assignment probability
defined above. Thus, \(U\) measures the proportion of the 44 assignments
that were not constant across replications. We also reported the three
most frequent full Q-matrix configurations to characterize how the
remaining replications were distributed among alternative matrices.

\subsubsection{Accuracy}\label{accuracy}

For the CoT workflow, we calculated accuracy across the 1,100 item-level
outputs per condition (11 items × 100 replications). Final answers were
compared with the answer key after normalization. Outputs without a
parseable answer field were treated as missing rather than incorrect.
Accuracy served as a task-performance check for interpreting the
reconstructed Q-matrices. It was not applicable to the direct workflow,
which generated Q-matrices without item-level answers. Details of the
normalization procedure are provided in the Supplementary Materials (see
Section S3.2.1).

\subsubsection{Model Fit}\label{model-fit}

We evaluated the empirical performance of the reference Q-matrix and the
20 consensus Q-matrices using responses from 536 examinees. All matrices
were fitted under the DINA model and the generalized DINA (G-DINA) model
(\citeproc{ref-delatorreGeneralizedDINAModel2011}{de la Torre, 2011}).
Under DINA, we compared model fit using \(-2\) log-likelihood
(\(-2\)LL), because all candidate Q-matrices involved the same number of
model parameters. Under G-DINA, we compared examinees' estimated
attribute profiles under each consensus Q-matrix with those estimated
under the reference Q-matrix. Profile agreement was defined as the
proportion of examinees whose profiles matched. As a separate check, we
evaluated the reference Q-matrix using the PVAF and Hull methods
reviewed in the Background.

\subsection{Results}\label{results-1}

\subsubsection{Alignment}\label{alignment-1}

\paragraph{Overall Overlap.}\label{overall-overlap}

Across conditions, the CoT-derived Q-matrices achieved an average
overlap of .809 with the reference Q-matrix, compared with .877 under
the direct workflow (see
\hyperref[tbl-empirical-performance-summary]{Table~\ref*{tbl-empirical-performance-summary}}),
a difference of .068. The exact item match rate, which requires all four
attributes of an item to match the reference, was .364 for the
CoT-derived Q-matrices and .564 under the direct workflow. These
differences correspond to approximately three of the 44 item--attribute
assignments and two of the 11 items per matrix, respectively. Across the
20 conditions, item-level overlap rates ranged from .50 to 1.00, and
Item 10 had an overlap rate of .50 in 10 of the 20 conditions.

The size and direction of the difference varied by model (see
\hyperref[fig-empirical-performance]{Figure~\ref*{fig-empirical-performance}}).
Kimi K2.6 achieved the highest overlap under the direct workflow (.932),
with exact matches for 8 of the 11 items. Its direct-workflow overlap
exceeded that under the CoT workflow by .091. DeepSeek v4 Flash also
achieved high overlap under the direct workflow (.909), with exact
matches for 7 of the 11 items at each of the three temperatures. Its
direct-workflow overlap was .091 higher than under the CoT workflow at
each temperature. Claude Haiku 4.5 showed the largest advantage for the
direct workflow: the overlap difference was .182 at temperatures 0 and
0.5, and .205 at temperature 1.0. GPT-4o mini showed the opposite
pattern and achieved higher overlap under the CoT workflow at every
temperature. Its overlap rates were .864, .886, and .886 at temperatures
0, 0.5, and 1.0, respectively, compared with .795 under the direct
workflow at each temperature.

\par\addvspace{\baselineskip}

\setcounter{table}{3}

\refstepcounter{table}%
\makeatletter\def\@currentlabel{4}\makeatother%
\label{tbl-empirical-performance-summary}

\apafloattitle{Table 4}

\apafloatcaption{Overall Overlap and Exact Item Match by Model,
Workflow, and Temperature}

\addcontentsline{lot}{table}{Table 4. Overall Overlap and Exact Item
Match by Model, Workflow, and Temperature}

{\def\LTcaptype{none} % do not increment counter
\begin{longtable}[]{@{}
  >{\raggedright\arraybackslash}p{(\linewidth - 8\tabcolsep) * \real{0.2400}}
  >{\raggedright\arraybackslash}p{(\linewidth - 8\tabcolsep) * \real{0.2200}}
  >{\raggedright\arraybackslash}p{(\linewidth - 8\tabcolsep) * \real{0.0900}}
  >{\raggedleft\arraybackslash}p{(\linewidth - 8\tabcolsep) * \real{0.2200}}
  >{\raggedleft\arraybackslash}p{(\linewidth - 8\tabcolsep) * \real{0.2300}}@{}}
\toprule\noalign{}
\begin{minipage}[b]{\linewidth}\raggedright
Model
\end{minipage} & \begin{minipage}[b]{\linewidth}\raggedright
Workflow
\end{minipage} & \begin{minipage}[b]{\linewidth}\raggedright
Temp.
\end{minipage} & \begin{minipage}[b]{\linewidth}\raggedleft
Overall overlap
\end{minipage} & \begin{minipage}[b]{\linewidth}\raggedleft
Exact item match
\end{minipage} \\
\midrule\noalign{}
\endhead
\bottomrule\noalign{}
\endlastfoot
GPT-4o mini & CoT workflow & 0 & \textbf{.864} & \textbf{.455} \\
& & 0.5 & \textbf{.886} & \textbf{.545} \\
& & 1 & \textbf{.886} & \textbf{.545} \\
& Direct workflow & 0 & .795 & .364 \\
& & 0.5 & .795 & .364 \\
& & 1 & .795 & .364 \\
DeepSeek v4 Flash & CoT workflow & 0 & .818 & .364 \\
& & 0.5 & .818 & .364 \\
& & 1 & .818 & .364 \\
& Direct workflow & 0 & .909 & .636 \\
& & 0.5 & .909 & .636 \\
& & 1 & .909 & .636 \\
Claude Haiku 4.5 & CoT workflow & 0 & .727 & .182 \\
& & 0.5 & .727 & .091 \\
& & 1 & .705 & .182 \\
& Direct workflow & 0 & .909 & .636 \\
& & 0.5 & .909 & .636 \\
& & 1 & .909 & .636 \\
Kimi K2.6 & CoT workflow & 0.6 & .841 & .545 \\
& Direct workflow & 0.6 & .932 & .727 \\
\end{longtable}
}

\apatablenoteskip

\begin{apafloatnote}

\emph{Note}. Boldface~values indicate conditions in which the CoT
workflow exceeded the direct workflow for the same model and
temperature.

\end{apafloatnote}

\par\addvspace{\baselineskip}

\begin{figure}[H]

\setcounter{figure}{3}

\refstepcounter{figure}%
\makeatletter\def\@currentlabel{4}\makeatother%
\label{fig-empirical-performance}

\apafloattitle{Figure 4}

\apafloatcaption{Overall Overlap by Model, Workflow, and Temperature}

\apafigureskip

\addcontentsline{lof}{figure}{Figure 4. Overall Overlap by Model,
Workflow, and Temperature}

\includegraphics[width=1\linewidth,height=\textheight,keepaspectratio]{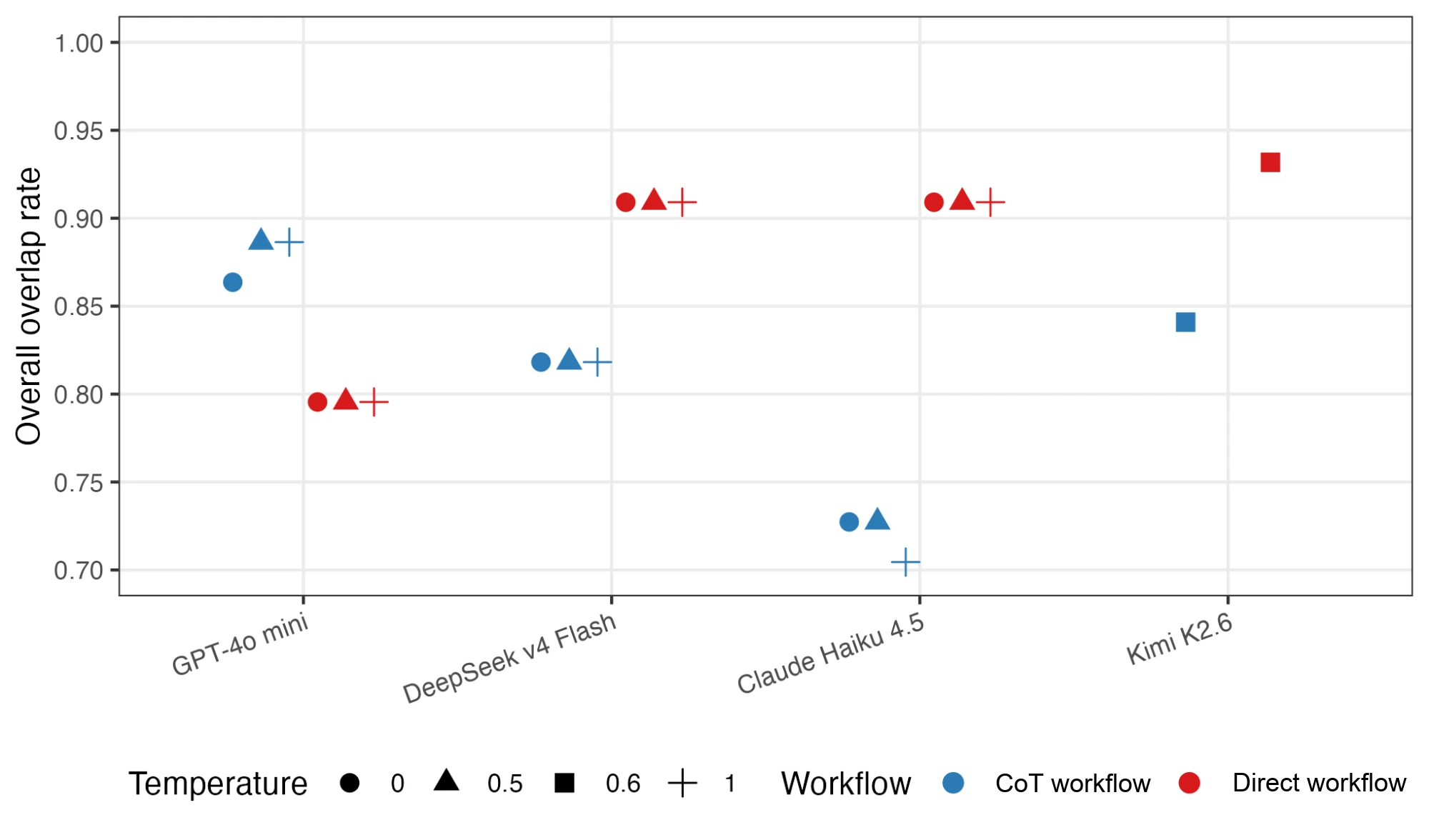}

\end{figure}

\FloatBarrier

\paragraph{Attribute-Level Overlap.}\label{attribute-level-overlap}

A1 (basic fraction subtraction) was the easiest attribute to recover
(see
\hyperref[tbl-empirical-attribute-summary]{Table~\ref*{tbl-empirical-attribute-summary}}).
All direct-workflow conditions achieved an A1 overlap rate of 1.000.
Under the CoT workflow, A1 overlap was also high, except for DeepSeek v4
Flash and Claude Haiku 4.5, both of which had rates of .909 across their
evaluated temperatures.

The remaining attribute-level results differed by workflow. Under the
CoT workflow, overlap losses were concentrated in A2
(simplifying/reducing) and A3 (separating whole number from fraction),
with A3 particularly weak for Claude Haiku 4.5. By contrast, A4
(borrowing from whole number) was recovered perfectly by GPT-4o mini and
DeepSeek v4 Flash under this workflow. Under the direct workflow, A3 was
recovered perfectly by every model except GPT-4o mini, which had an
overlap rate of .727. A4 was also recovered perfectly by every model
except GPT-4o mini (.636--.727) and Kimi K2.6 (.818, on Items 1 and 5).
A2 was the weakest attribute for DeepSeek v4 Flash and Claude Haiku 4.5
at every temperature. These results indicate that LLMs recovered A1
consistently but varied in recovering the other attributes, which
underscores the importance of evaluating recovery at the attribute
level.

\par\addvspace{\baselineskip}

\setcounter{table}{4}

\refstepcounter{table}%
\makeatletter\def\@currentlabel{5}\makeatother%
\label{tbl-empirical-attribute-summary}

\apafloattitle{Table 5}

\apafloatcaption{Attribute-Level Overlap by Model, Workflow, and
Temperature}

\addcontentsline{lot}{table}{Table 5. Attribute-Level Overlap by Model,
Workflow, and Temperature}

{\def\LTcaptype{none} % do not increment counter
\begin{longtable}[]{@{}
  >{\raggedright\arraybackslash}p{(\linewidth - 12\tabcolsep) * \real{0.2400}}
  >{\raggedright\arraybackslash}p{(\linewidth - 12\tabcolsep) * \real{0.2600}}
  >{\raggedright\arraybackslash}p{(\linewidth - 12\tabcolsep) * \real{0.0800}}
  >{\raggedleft\arraybackslash}p{(\linewidth - 12\tabcolsep) * \real{0.1000}}
  >{\raggedleft\arraybackslash}p{(\linewidth - 12\tabcolsep) * \real{0.1000}}
  >{\raggedleft\arraybackslash}p{(\linewidth - 12\tabcolsep) * \real{0.1100}}
  >{\raggedleft\arraybackslash}p{(\linewidth - 12\tabcolsep) * \real{0.1100}}@{}}
\toprule\noalign{}
\begin{minipage}[b]{\linewidth}\raggedright
Model
\end{minipage} & \begin{minipage}[b]{\linewidth}\raggedright
Workflow
\end{minipage} & \begin{minipage}[b]{\linewidth}\raggedright
Temp.
\end{minipage} & \begin{minipage}[b]{\linewidth}\raggedleft
A1
\end{minipage} & \begin{minipage}[b]{\linewidth}\raggedleft
A2
\end{minipage} & \begin{minipage}[b]{\linewidth}\raggedleft
A3
\end{minipage} & \begin{minipage}[b]{\linewidth}\raggedleft
A4
\end{minipage} \\
\midrule\noalign{}
\endhead
\bottomrule\noalign{}
\endlastfoot
GPT-4o mini & CoT workflow & 0 & 1.000 & .455 & 1.000 & 1.000 \\
& & 0.5 & 1.000 & .545 & 1.000 & 1.000 \\
& & 1 & 1.000 & .636 & .909 & 1.000 \\
& Direct workflow & 0 & 1.000 & .818 & .727 & .636 \\
& & 0.5 & 1.000 & .727 & .727 & .727 \\
& & 1 & 1.000 & .727 & .727 & .727 \\
DeepSeek v4 Flash & CoT workflow & 0 & .909 & .636 & .727 & 1.000 \\
& & 0.5 & .909 & .636 & .727 & 1.000 \\
& & 1 & .909 & .636 & .727 & 1.000 \\
& Direct workflow & 0 & 1.000 & .636 & 1.000 & 1.000 \\
& & 0.5 & 1.000 & .636 & 1.000 & 1.000 \\
& & 1 & 1.000 & .636 & 1.000 & 1.000 \\
Claude Haiku 4.5 & CoT workflow & 0 & .909 & .636 & .545 & .818 \\
& & 0.5 & .909 & .636 & .455 & .909 \\
& & 1 & .909 & .636 & .364 & .909 \\
& Direct workflow & 0 & 1.000 & .636 & 1.000 & 1.000 \\
& & 0.5 & 1.000 & .636 & 1.000 & 1.000 \\
& & 1 & 1.000 & .636 & 1.000 & 1.000 \\
Kimi K2.6 & CoT workflow & 0.6 & 1.000 & .636 & .818 & .909 \\
& Direct workflow & 0.6 & 1.000 & .909 & 1.000 & .818 \\
\end{longtable}
}

\par\addvspace{\baselineskip}

\FloatBarrier

\paragraph{Direction and Location of
Disagreements.}\label{direction-and-location-of-disagreements}

In the four CoT workflow conditions examined for disagreement patterns,
32 item--attribute assignments in the consensus Q-matrices disagreed
with the reference Q-matrix. Of these, 29 were omissions and 3 were
over-labelings. Eighteen of the 29 omissions were concentrated in five
item--attribute locations. A2 was omitted on Items 1, 5, 10, and 11 in
all four conditions, which accounts for 16 disagreements. On Item 11,
the assignment probabilities for A2 ranged from .00 to .31. A1 was
omitted on Item 3 by DeepSeek v4 Flash and Claude Haiku 4.5, which
accounts for two more disagreements. The remaining 11 omissions occurred
mainly on A3 for Claude Haiku 4.5 across five items, with fewer A3
omissions for DeepSeek v4 Flash and Kimi K2.6 (three and one,
respectively). They also included A4 on Item 1 for Claude Haiku 4.5 and
Kimi K2.6.

The three over-labelings were A3 on Item 6 for Claude Haiku 4.5 and Kimi
K2.6, and A2 on Item 2 for GPT-4o mini. The reference Q-matrix had a
density of .727. All four consensus Q-matrices in these conditions were
sparser: Claude Haiku 4.5 had a density of .500, DeepSeek v4 Flash .545,
Kimi K2.6 .614, and GPT-4o mini .659.

\subsubsection{Structural
Identifiability}\label{structural-identifiability-1}

None of the 2,000 generated Q-matrices or the 20 consensus Q-matrices
satisfied all three DINA identifiability conditions. All failed the
completeness condition because none contained single-attribute items for
all four attributes. The reference Q-matrix also failed completeness,
although it satisfied the repeated-measurement and distinct-column
conditions. Specifically, it contained an A1-only item but no
single-attribute items for A2, A3, or A4, because every item requiring
one of these attributes also required A1. The two workflows therefore
did not differ on this criterion.

\subsubsection{Stability}\label{stability-1}

The direct workflow generally produced more stable Q-matrices than the
CoT workflow (see
\hyperref[tbl-empirical-variation-summary]{Table~\ref*{tbl-empirical-variation-summary}}
and
\hyperref[fig-empirical-variation]{Figure~\ref*{fig-empirical-variation}}).
The mean unstable item--attribute assignment rate (\(U\)) was .248 for
the direct workflow and .409 for the CoT workflow, and \(U\) was lower
for the direct workflow in nine of the 10 model--temperature comparisons
and equal in one (GPT-4o mini at temperature 1.0). The mean dominant
Q-matrix proportion (\(D\)) was .540 and .044, respectively, and was
higher for the direct workflow in every comparison, although \(D\) is
not directly comparable across workflows because the two workflows
generate complete matrices differently.

The three most frequent full Q-matrix configurations accounted for
.15--1.00 of replications in the direct workflow, compared with .06--.25
in the CoT workflow (see Table S3). Within the direct workflow, \(D\)
decreased with temperature for GPT-4o mini (.45 to .06) and Claude Haiku
4.5 (1.00 to .18), whereas it remained between .60 and .74 for DeepSeek
v4 Flash. For Kimi K2.6, \(D\) was .50, and its three most frequent
configurations accounted for .94 of replications.

\par\addvspace{\baselineskip}

\setcounter{table}{5}

\refstepcounter{table}%
\makeatletter\def\@currentlabel{6}\makeatother%
\label{tbl-empirical-variation-summary}

\apafloattitle{Table 6}

\apafloatcaption{Empirical Q-Matrix Stability and Accuracy by Model,
Workflow, and Temperature}

\addcontentsline{lot}{table}{Table 6. Empirical Q-Matrix Stability and
Accuracy by Model, Workflow, and Temperature}

{\def\LTcaptype{none} % do not increment counter
\begin{longtable}[]{@{}
  >{\raggedright\arraybackslash}p{(\linewidth - 10\tabcolsep) * \real{0.2400}}
  >{\raggedright\arraybackslash}p{(\linewidth - 10\tabcolsep) * \real{0.2600}}
  >{\raggedright\arraybackslash}p{(\linewidth - 10\tabcolsep) * \real{0.0800}}
  >{\raggedleft\arraybackslash}p{(\linewidth - 10\tabcolsep) * \real{0.1400}}
  >{\raggedleft\arraybackslash}p{(\linewidth - 10\tabcolsep) * \real{0.1400}}
  >{\raggedleft\arraybackslash}p{(\linewidth - 10\tabcolsep) * \real{0.1400}}@{}}
\toprule\noalign{}
\begin{minipage}[b]{\linewidth}\raggedright
Model
\end{minipage} & \begin{minipage}[b]{\linewidth}\raggedright
Workflow
\end{minipage} & \begin{minipage}[b]{\linewidth}\raggedright
Temp.
\end{minipage} & \begin{minipage}[b]{\linewidth}\raggedleft
\(D\)
\end{minipage} & \begin{minipage}[b]{\linewidth}\raggedleft
\(U\)
\end{minipage} & \begin{minipage}[b]{\linewidth}\raggedleft
Accuracy
\end{minipage} \\
\midrule\noalign{}
\endhead
\bottomrule\noalign{}
\endlastfoot
GPT-4o mini & CoT workflow & 0 & .10 & .250 & .738 \\
& & 0.5 & .04 & .409 & .770 \\
& & 1 & .03 & .455 & .787 \\
& Direct workflow & 0 & .45 & .205 & - \\
& & 0.5 & .14 & .295 & - \\
& & 1 & .06 & .455 & - \\
DeepSeek v4 Flash & CoT workflow & 0 & .02 & .591 & .998 \\
& & 0.5 & .02 & .432 & .999 \\
& & 1 & .03 & .432 & .997 \\
& Direct workflow & 0 & .74 & .318 & - \\
& & 0.5 & .60 & .295 & - \\
& & 1 & .73 & .341 & - \\
Claude Haiku 4.5 & CoT workflow & 0 & .09 & .227 & .999 \\
& & 0.5 & .03 & .409 & .980 \\
& & 1 & .04 & .409 & .996 \\
& Direct workflow & 0 & 1.00 & .000 & - \\
& & 0.5 & 1.00 & .000 & - \\
& & 1 & .18 & .341 & - \\
Kimi K2.6 & CoT workflow & 0.6 & .04 & .477 & .995 \\
& Direct workflow & 0.6 & .50 & .227 & - \\
\end{longtable}
}

\apatablenoteskip

\begin{apafloatnote}

\emph{Note}. \(D\)~= dominant Q-matrix proportion. \(U\) = unstable
item--attribute assignment rate. A hyphen indicates not applicable.

\end{apafloatnote}

\par\addvspace{\baselineskip}

\begin{figure}[H]

\setcounter{figure}{4}

\refstepcounter{figure}%
\makeatletter\def\@currentlabel{5}\makeatother%
\label{fig-empirical-variation}

\apafloattitle{Figure 5}

\apafloatcaption{Q-Matrix Stability by Model and Temperature}

\apafigureskip

\addcontentsline{lof}{figure}{Figure 5. Q-Matrix Stability by Model and
Temperature}

\includegraphics[width=1\linewidth,height=\textheight,keepaspectratio]{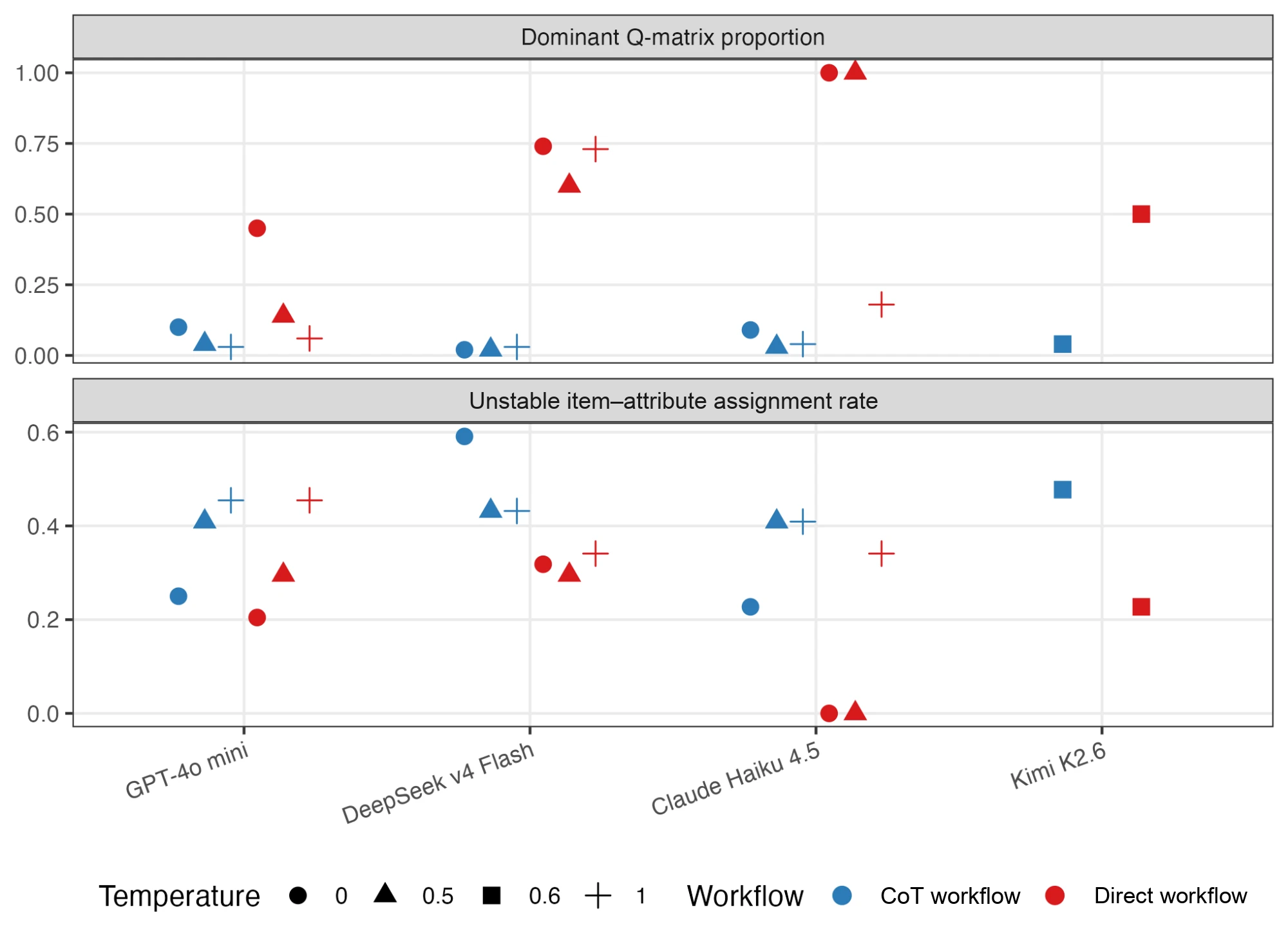}

\end{figure}

\FloatBarrier

\subsubsection{Accuracy}\label{accuracy-1}

Under the CoT workflow, DeepSeek v4 Flash, Claude Haiku 4.5, and Kimi
K2.6 each achieved accuracy of at least .98 across all conditions. Their
\(D\) values ranged from .02 to .09, while their \(U\) values ranged
from .227 to .591. GPT-4o mini had lower accuracy (.738 to .787 across
temperatures). Because the three models with near-perfect accuracy also
produced unstable Q-matrices, instability was not explained by incorrect
answers.

\subsubsection{Model Fit}\label{model-fit-1}

Across the 536 examinees, DINA fit varied among Q-matrices with similar
levels of overlap with the reference. The DeepSeek v4 Flash CoT-derived
Q-matrix had a slightly lower \(-2\)LL than the reference (5044.9
vs.~5045.8), despite an overall overlap rate of .818. The GPT-4o mini
direct-workflow Q-matrix had the worst fit, with a \(-2\)LL 347 points
higher than the reference, despite an overlap rate of .795. Of the 20
consensus Q-matrices, 13 were distinct. Across these 13 matrices,
overlap and \(-2\)LL had a moderate negative association that was not
statistically significant (Spearman \(\rho = -.49\), \(p = .09\), see
Table S4).

Under G-DINA, 64\% of examinees received the same estimated attribute
profile under the DeepSeek v4 Flash CoT-derived Q-matrix as under the
reference Q-matrix. Both the PVAF and Hull methods flagged A2 on Items 1
and 5 and A4 on Item 10 in the reference Q-matrix as candidates for
changing from 1 to 0 (see Table S5, which also reports a third
validation method). A2 on Items 1 and 5 were also among the cells the
LLMs departed from most often, with A2 on Item 1 omitted in 16 of the 20
consensus Q-matrices and A2 on Item 5 in 18, across both workflows.

\section{Discussion}\label{discussion}

This study proposed RTA as a framework for examining the reasoning
behind LLM-generated item--attribute assignments and evaluated it as a
supplement to Q-matrix specification. When an LLM acts as an expert in
Q-matrix construction, responsible use requires that the basis of its
assignments can be inspected
(\citeproc{ref-johnsonResponsibleAIMeasurement2025}{Johnson, 2025}), and
the validity and reliability of the resulting diagnostic classifications
depend on the Q-matrix it produces. The simulation study showed that the
CoT workflow recovered the required attributes under controlled
conditions, and the empirical study showed that the resulting
CoT-derived Q-matrices overlapped substantially with the reference
Q-matrix (mean overlap .809) but diverged from it at specific
item--attribute locations. Because these results draw on a comprehensive
evaluation of metrics from several perspectives, the sections below
interpret them by research question, addressing what the recovered
traces reveal, why the matrices diverge from the reference, and how
stable they are, before turning to the limitations and implications of
the study.

\subsection{Reasoning-Process Recovery
(RQ1)}\label{reasoning-process-recovery-rq1}

The results generally support H1 at the attribute level. The CoT
workflow recovered nearly all expected attributes under controlled
perturbations, and the variants with absent attributes showed that this
recovery did not depend on assigning all four attributes by default.
However, the results do not show recovery of a unique ordered reasoning
process. Different step sequences can express the same set of
attributes, consistent with treating elicited traces as structured
material for examining problem-solving activity
(\citeproc{ref-liUnderstandingThinkingProcess2025}{M. Li, Zhang, et al.,
2025}). GPT-4o mini's subtract-then-borrow route is one example: it
could reflect case-based reasoning in the sense of Hu et al.
(\citeproc{ref-huCaseBasedRuleBased2024}{2024}), but attempting
subtraction before borrowing is also a legitimate problem-solving
heuristic.

A more direct contribution of RTA is that trace inspection shows where
an expressed operation and its skill label diverge, which cannot be seen
from the final matrix alone. A reconstructed Q-vector records only which
labels appeared, so an omitted attribute may mean that the operation was
not performed or that it was performed inside a step carrying a
different label. Reading the traces showed that several omissions
reflected the second case: models carried out the operation but folded
it into an adjacent step labeled with another attribute. An alternative
explanation is that omissions arise when the model answers the item
incorrectly and therefore never carries out the operation. To examine
this, a post hoc analysis re-derived the consensus Q-matrix of GPT-4o
mini from answer-correct responses only (see Supplementary Experiment 1,
Section S4.1). The assignments did not change, which indicates that, for
this model, label omission does not stem from arithmetic errors.
Accuracy alone therefore does not indicate the quality of a
reconstructed Q-matrix, and inspecting the expressed steps adds
information beyond the final matrix. RTA makes visible differences in
how models solve items that accuracy scores conceal. In measurement
terms, these traces provide evidence based on response processes for the
generated assignments
(\citeproc{ref-aeraStandardsEducationalPsychological2014}{American
Educational Research Association et al., 2014}), which matrix-level
metrics lack.

\subsection{LLM--Human Alignment (RQ2)}\label{llmhuman-alignment-rq2}

Both workflows achieved substantial overlap with the reference Q-matrix,
consistent with prior evidence that LLMs can approximate
expert-reference matrices at the aggregate level
(\citeproc{ref-asiretEvaluatingChatGPT4osPerformance2025}{Aşiret \& Ömür
Sünbül, 2025};
\citeproc{ref-xueEvaluatingGeneralPurposeMultimodal2026}{Xue \&
Appleton, 2026}). This supports the first part of H2. However, the
expected comparability between workflows was not supported: the direct
workflow achieved higher overlap for three of the four models, whereas
GPT-4o mini showed the opposite pattern.

Reference agreement alone does not show that the required operations
were correctly identified. The reference Q-matrix is an expert
specification rather than a ground truth, as the PVAF and Hull methods
flagged three of its cells as candidates for removal. The DeepSeek v4
Flash CoT-derived Q-matrix diverged from the reference yet fit the
response data comparably and assigned different attribute profiles to a
substantial share of examinees, so no single index of overlap, fit, or
profile agreement validates a Q-matrix on its own.

The direct workflow presents a further interpretive problem because it
provides no intermediate output to inspect. An assignment that matches
the reference could reflect an accurate assessment of the item's
required operations or retrieval of a familiar matrix encountered during
pretraining. This alternative deserves consideration because the item
set and its reference Q-matrix have been published and reused since
(\citeproc{ref-orenProvingTestSet2024}{Oren et al., 2024}; Tatsuoka;
\citeproc{ref-tatsuoka1990}{1990}). Because no trace is available to
separate the two, the direct workflow was re-run on isomorphic items,
which preserve each item's structure but change its numerical values
(see Supplementary Experiment 2, Section S4.2). Alignment largely
persisted, so retrieval does not explain the direct workflow's alignment
advantage in general, although it remains possible for Claude Haiku 4.5
at low temperatures.

The direction of disagreement further illustrates what aggregate overlap
can conceal. For the four models under the CoT workflow, omissions
substantially outnumbered over-labelings, and all four CoT-derived
consensus matrices were sparser than the reference, consistent with the
density divergence reported by Xue and Appleton
(\citeproc{ref-xueEvaluatingGeneralPurposeMultimodal2026}{2026}). The
asymmetry also partly reflects the reconstruction rule noted under RQ1:
because an attribute is included only when a step receives its label, an
operation expressed within a step labeled with a different attribute is
missed, whereas over-labeling requires a step to carry an attribute that
the reference does not assign to the item.

\subsubsection{Sources of LLM--Human
Disagreement}\label{sources-of-llmhuman-disagreement}

A disagreement with the reference should prompt further inspection
rather than be treated automatically as an error in the generated matrix
or its reasoning trace, because the two matrices record different
things. The reference Q-matrix reflects an expert's classification of
each item, whereas the CoT-derived Q-matrix records the attributes
expressed through a particular solution trace. Inspecting the traces
behind the omissions and over-labelings showed that the disagreements
arose from four sources, which a matrix-level comparison alone would not
have separated.

Two sources lay in the specification rather than in the model's
reasoning. The reference assigns attributes by item type, whereas a
trace reflects the operations that a particular numerical instance
requires, so an assigned attribute may be unnecessary for the instance
actually solved. In addition, the prompt's labeling instruction was
narrower than the attribute definition, which led models to label
already-simplified results as {[}Other{]}. The direct workflow and the
PVAF and Hull analyses showed some of the same disagreements, so they
were not specific to CoT labeling. Removing the labeling rules for
Claude Haiku 4.5 reversed most of the affected assignments (see
Supplementary Experiment 3, Section S4.3), so the prompt that elicits
and labels the steps is part of the RTA specification.

The other two sources lay in the trace itself. An operation could be
performed without being represented as a distinct, labelable step, so
attribute recovery depends on how an operation is expressed as well as
on whether it is performed. Conversely, a step outside the item's
required operations, such as converting an improper fraction to a mixed
number, could receive an attribute label. Constraining the final answer
to mixed-number form raised this over-labeling without changing accuracy
(see Supplementary Experiment 4, Section S4.4), which suggests that the
label responded to answer format rather than to the solution itself.

\subsection{Q-Matrix Stability and Variation
(RQ3)}\label{q-matrix-stability-and-variation-rq3}

The results generally support H3, particularly at the item--attribute
level. Because \(D\) is not directly comparable across workflows (see
Stability in the Empirical Study), cross-workflow conclusions rely
primarily on \(U\). Both workflows produced varied Q-matrices across
replications, and accuracy did not account for this variation, since
models that answered almost all items correctly under the CoT workflow
still produced diverse matrices. Run-to-run variation in LLM-generated
Q-matrices has also been reported by Kim
(\citeproc{ref-kimHowReproducibleAre2026}{2026}), who found that
regenerating a Q-matrix with the same model at the same temperature
changed the mastery profiles assigned to many examinees. A single run
therefore yields only one draw from a distribution of possible matrices.
Because diagnostic classifications inherit any instability in the
Q-matrix, this variation bears on the reliability of classifications
based on a single generated Q-matrix. The direct workflow's greater
stability could reflect retrieval of a familiar matrix (see RQ2), but
stability alone does not show this mechanism or imply correctness.
Claude Haiku 4.5 illustrates this: it repeatedly produced the same
matrix at temperatures 0 and 0.5, yet that matrix still disagreed with
the reference, so consistency across replications did not indicate
agreement with it.

\subsection{Limitations and Future
Directions}\label{limitations-and-future-directions}

Several limitations qualify these findings. First, the study used one
fraction-subtraction item set and reference Q-matrix, so the findings
may not extend beyond this domain. Because every item requiring A2, A3,
or A4 also required A1, none of the matrices, including the reference,
satisfied the completeness condition for DINA identifiability. Second,
RTA analyzes elicited text rather than internal computation, and the
model-generated skill labels were not independently coded. Third, the
simulation and empirical studies differed in attribute wording and in
whether a worked example was included, so their results are not directly
comparable. Fourth, the workflows differed in call structure (11
item-level calls versus one full-matrix response), which may have
contributed to their stability differences. Fifth, three of the four
models were lightweight variants, and agreement with a validated
Q-matrix can change across model versions
(\citeproc{ref-fanUseAIToolsQMatrix2026}{Fan et al., 2026}), so the
results may not hold for other models or later versions. Finally, Kimi
K2.6 was evaluated at only one temperature, and Claude Haiku 4.5's
reasoning setting took effect only at temperature 1.0, which confounds
temperature with reasoning effort (see Supplementary Experiment 5,
Section S4.5).

Future work should evaluate RTA on newly developed item sets and in
other domains, validate trace coding through independent expert review,
and systematically vary prompt wording, worked examples, and call
structure. Trace analysis should also be combined with response-data
validation, while item-pool identifiability should be assessed before
fitting a DCM.

\subsection{Implications}\label{implications}

RTA contributes inspectable evidence about how individual
item--attribute assignments are expressed and where disagreements arise,
which matrix alignment and stability alone do not provide. Unlike a
rationale that accompanies a Q-vector
(\citeproc{ref-zhangCDMQMatrixDiscovery2026}{Zhang \& Rao, 2026}), an
RTA trace produces the Q-vector, so each assignment can be traced to the
step that generated it. Three practices follow from the results. The
prompt that elicits and labels the steps should be reported with the
results, generated Q-matrices should be summarized across replications
rather than read from a single run, and disagreements with the reference
should be inspected before they are treated as errors. Because elicited
traces are prompt-dependent and do not directly reveal internal
computation, RTA is best used to complement expert judgment,
response-data validation, and structural identifiability checks. In this
role, RTA supports the transparency and human oversight that the
responsible use of AI in measurement requires
(\citeproc{ref-johnsonResponsibleAIMeasurement2025}{Johnson, 2025}),
because each generated assignment can be traced, reviewed, and
contested.

\FloatBarrier

\section{References}\label{references}

\protect\phantomsection\label{refs}

\end{document}